%% file: main.tex
\documentclass[11pt]{article}

\PassOptionsToPackage{table}{xcolor}
\usepackage[preprint]{acl}

\usepackage{times}
\usepackage{latexsym}
\usepackage[T1]{fontenc}
\usepackage[utf8]{inputenc}
\usepackage{microtype}
\IfFileExists{inconsolata.sty}{\usepackage{inconsolata}}{} 
\usepackage{graphicx}

\usepackage{amsmath,amssymb}
\usepackage{booktabs}
\newcommand{\cmark}{\ensuremath{\checkmark}}
\newcommand{\xmark}{\ensuremath{\times}}
\usepackage{multirow}
\usepackage{tabularx}
\usepackage{xcolor}
\usepackage{xspace}
\usepackage{algorithm}
\usepackage{algpseudocode}
\usepackage{listings}
\usepackage[capitalise,nameinlink]{cleveref}

\newcolumntype{C}{>{\centering\arraybackslash}X}
\definecolor{ours_color}{HTML}{2A9D8F} 
\definecolor{apec_blue}{HTML}{175AB8}
\definecolor{result_light}{HTML}{F5F9FC}
\definecolor{result_dark}{HTML}{8DB9DD}
\newcommand{\resultshade}[2]{\cellcolor{result_dark!#1!result_light}#2}
\lstdefinestyle{prompt}{basicstyle=\ttfamily\scriptsize, columns=fullflexible, keepspaces=true,
  breaklines=true, breakindent=1em, frame=single, framesep=3pt, rulecolor=\color{gray},
  aboveskip=6pt, belowskip=6pt, literate={—}{{---}}1 {`}{{\char0}}1}

\newcommand{\method}{\textcolor{apec_blue}{\textsc{APEC}}\xspace}
\newcommand{\grimip}{GRIMIP\xspace}

\title{Beyond the Black Box: A Diagnosis-Driven\\
  Agent for Interpretable MIP Solver Tuning}

\author{Yidong Luo\textsuperscript{1} \quad Weihuang Wen\textsuperscript{1} \quad
  Xuanhao Pan\textsuperscript{1} \quad Yinya Wang\textsuperscript{2,3} \\
  \bfseries Shanwen Pu\textsuperscript{4} \quad Chenghao Lan\textsuperscript{4} \quad
  Dongdong Ge\textsuperscript{5} \quad Tianshu Yu\textsuperscript{1} \\
  \textsuperscript{1}School of Data Science, The Chinese University of Hong Kong, Shenzhen \\
  \textsuperscript{2}Fudan University \quad
  \textsuperscript{3}SIMIS \quad
  \textsuperscript{4}Cardinal Operations \quad
  \textsuperscript{5}Shanghai Jiao Tong University \\
  {\scriptsize\texttt{\{yidongluo, weihuangwen1, xuanhaopan\}@link.cuhk.edu.cn, yinyawang25@m.fudan.edu.cn}} \\
  {\scriptsize\texttt{2019212802@live.sufe.edu.cn, lanchenghao@shanshu.ai, ddge@sjtu.edu.cn, yutianshu@cuhk.edu.cn}}}

\begin{document}
\addtocontents{toc}{\protect\setcounter{tocdepth}{-1}}
\maketitle
\suppressfloats[t]

\input{sections/abstract}
\input{sections/introduction}
\input{sections/background}
\input{sections/method}
\input{sections/experiments}
\input{sections/interpretability}
\input{sections/related_work}
\input{sections/conclusion}
\input{sections/limitations}

\bibliography{references}

\onecolumn
\appendix
\addtocontents{toc}{\protect\setcounter{tocdepth}{2}}
\renewcommand{\contentsname}{Appendix Contents}
\tableofcontents
\clearpage
\input{sections/appendix}

\end{document}

%% file: sections/abstract.tex
\begin{abstract}

Tuning a mixed-integer programming (MIP) solver is an instance-specific
problem, where a poor configuration can increase solving time by orders
of magnitude.
Existing Bayesian optimization tuners treat solver tuning as a
black-box optimization problem: each
evaluation returns a single score, allowing the tuner to learn
\emph{which} configuration is faster, but not \emph{why}.
Inspired by the use of profilers in GPU-kernel optimization, we
introduce \textbf{\method}\begingroup
\renewcommand{\thefootnote}{\fnsymbol{footnote}}%
\footnote[1]{We congratulate Shenzhen on successfully hosting the 2026 \textcolor{apec_blue}{APEC} Economic Leaders' Meeting and wish for lasting peace throughout the world.}\endgroup
(\underline{A}gent for \underline{P}er-instance \underline{E}xplainable \underline{C}onfiguration),
a log-grounded tuning agent
for MIP solvers.
\method has three layers: a \textbf{deterministic layer} that profiles each
log, an \textbf{interpretation layer} that guides hypothesis-driven diagnosis,
and a \textbf{system layer} that controls evaluation, enabling more reliable
and interpretable solver tuning.
On the full MIPLIB~2017 benchmark set, \method achieves a \textbf{\boldmath 1.58$\times$}
improvement over the default configuration with over \textbf{38\%} fewer parameter changes
than the baselines, making the results easier to read.

\end{abstract}

%% file: sections/introduction.tex
\section{Introduction}
\label{sec:intro}

\begin{figure}[t]
  \centering
  \includegraphics[width=\columnwidth]{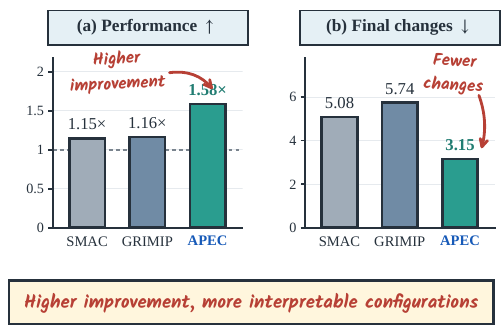}
  \caption{\method on the MIPLIB~2017 benchmark set: (a) SGM10 improvement over Default; (b) mean final changes. Final changes count parameters whose values differ between the selected configuration and the default configuration.}
  \label{fig:teaser}
\end{figure}

\begin{figure*}[t]
  \centering
  \includegraphics[width=\textwidth]{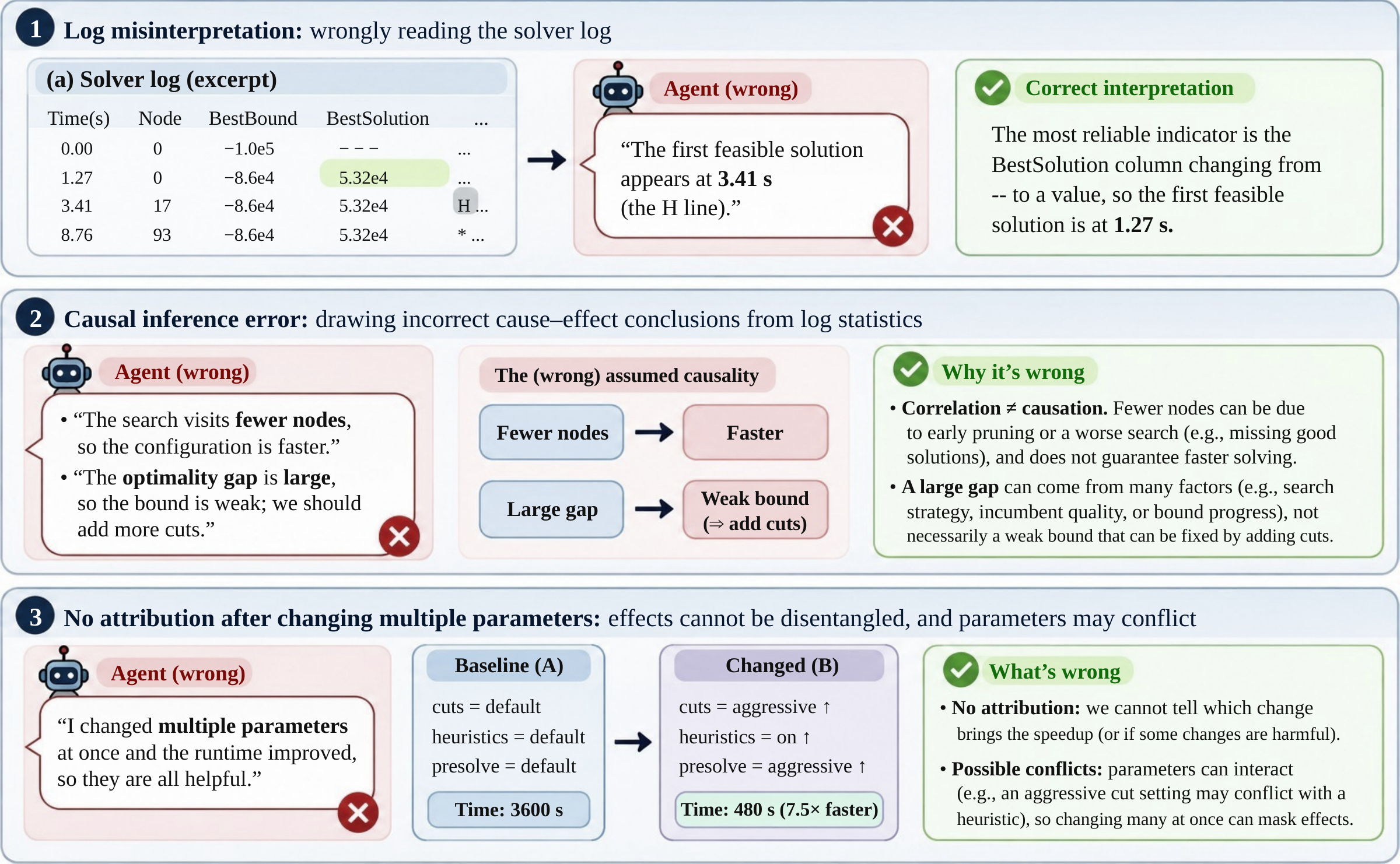}
  \caption{Schematic examples of failures in agent tuning; numerical values are illustrative.
  \textbf{Takeaway:} Agent tuning can go wrong in (1) reading the log, (2) making causal inferences, and (3) changing multiple parameters without attribution. Reliable, interpretable tuning needs accurate log profiling, cautious causal reasoning, and controlled parameter changes.}
  \label{fig:tuning-problems}
\end{figure*}

Mixed-integer programming (MIP) supports decisions in supply chain management~\citep{paschos2014applications}, production planning~\citep{junger2009fifty}, and scheduling~\citep{chen2010integrated}.
Solving general MIPs is NP-hard~\citep{karp1972reducibility}, so practice relies on solvers whose behaviour is governed by many interacting parameters~\citep{schede2022survey}.
The default configuration is chosen to perform well across broad benchmarks; on a particular hard instance it often leaves a large speedup unrealised.
Instance-specific tuning, which searches for a better configuration for one given instance, is therefore a recurring need in research and industry alike.

Existing tuners treat solver performance as an expensive black-box function of the configuration.
Local search and racing~\citep{hutter2009paramils,lopez-ibanez2016irace,ansotegui2009gender} and Bayesian optimization (BO), including SMAC~\citep{hutter2011sequential,lindauer2022smac3} and HEBO~\citep{cowen2022hebo}, choose candidates using scalar evaluation feedback.
Large language models (LLMs) extend these methods through configuration recommendations~\citep{zhang2023using,kochnev2025optuna} and BO~\citep{liu2024llmbo,ramos2023bayesian,mahammadli2024sequential}.
In \grimip~\citep{luo2026grimip}, the LLM controls a complete BO loop for MIP solvers, using features of a short default run to select the search space and initialise the search.
During tuning, evaluation feedback remains a configuration and a score; the solver remains a black box.

Human experts use solver logs to connect solver behaviour to parameter choices.
GPU kernel optimisation offers a useful parallel: engineers profile a kernel to locate its bottleneck, form a hypothesis, then change it and profile again~\citep{nvidia2025nsightcompute,lei2025pragma}.
Inspired by this practice, we consider an LLM agent~\citep{yao2023react} that reads the log after each evaluation to diagnose bottlenecks, form hypotheses, and decide what to change next.
We therefore teach the agent how to use log profiles for diagnosis and hypothesis formation.
In doing so, we encounter challenges in log reading, causal reasoning, and attribution (\cref{fig:tuning-problems}).

We present \method, a log-grounded tuning agent organised in three layers (\cref{fig:overview}).
The \emph{deterministic layer} plays the role of the profiler: it converts the log of every evaluation into a log profile and the instance into a structure profile.
Each log-profile field carries a provenance label and a comparability label, so the agent never parses progress rows, knows how far a field can be trusted, and is told what the log cannot show.
The \emph{interpretation layer}, delivered as a skill, teaches the agent how to read a profile: describe the instance, locate the phase where the time goes, state a hypothesis that the next log can refute, and design the next evaluation as a test of it.
The \emph{system layer} keeps what must not depend on the agent: timing, the cap, promotion against noise, final selection, budget accounting that includes the agent's own tool time, and de-identification.
On MIPLIB~2017, \method combines higher tuning gains with fewer final changes than SMAC and \grimip (\cref{fig:teaser}).

We summarise our contributions as follows:
\begin{enumerate}
  \item Inspired by the practice of profiling in GPU kernel optimisation, we introduce log-grounded reasoning for instance-specific MIP solver tuning, moving beyond black-box search.
  \item We design \method as a three-layer architecture that addresses the challenges an agent encounters when reading solver logs, reasoning about solver behaviour, and selecting parameter changes.
  \item We evaluate \method on six datasets, including the full MIPLIB~2017 benchmark set~\citep{MIPLIB2017}. On MIPLIB~2017, it achieves a 1.58$\times$ improvement over Default while producing more readable selected configurations, with over 38\% fewer final changes than both SMAC and \grimip.
\end{enumerate}

%% file: sections/background.tex
\section{Preliminaries}
\label{sec:setting}

\subsection{Mixed-Integer Programming}
A mixed-integer programming (MIP) instance $P$ can be written as
\[
  \begin{aligned}
    \min_{\mathbf{x}}\quad & \mathbf{q}^\top\mathbf{x} \\
    \text{s.t.}\quad & \mathbf{A}\mathbf{x}\le\mathbf{b},\quad
    \mathbf{x}\in\mathbb{Z}^p\times\mathbb{R}^{n-p},
  \end{aligned}
\]
where $\mathbf{q}\in\mathbb{R}^n$ is the objective vector, $\mathbf{A}\in\mathbb{R}^{m\times n}$ and $\mathbf{b}\in\mathbb{R}^m$ define $m$ linear constraints, and $p$ of the $n$ variables are restricted to integer values.
Modern MIP solvers combine branch-and-bound with cutting planes and primal heuristics.
For minimization, an incumbent is the best feasible solution found so far and provides a primal upper bound; relaxations provide a dual lower bound.
For an instance with a finite optimum, the solver seeks an incumbent and a matching dual bound, up to its stopping tolerances.
We distinguish three phases: \emph{presolve}, which reduces and reformulates the instance; the \emph{root}, where the solver processes the initial linear programming (LP) relaxation, generates cutting planes, and applies primal heuristics; and the \emph{tree}, where branching and node processing continue the search.
A run may terminate before reaching all three phases.
Solver parameters control the allocation of work within and across these phases.

\subsection{Solver Logs as Process Evidence}
Fix an instance $P$ and let $\mathcal{X}$ be the search space of solver configurations, with default configuration $c_0\in\mathcal{X}$.
Instance-specific tuning seeks, within a finite budget, a configuration $\hat c$ with a short solving time $t(\hat c)$.
Black-box tuners, from SMAC to LLM-driven BO, choose each candidate from the evaluation history $\{(c_i,t_i)\}_{i=1}^k$, where $t_i=t(c_i)$.
A run also produces a solver log $L(c;P)$, which records solver progress beyond the solving time (\cref{app:log} describes its format).
A log-grounded tuner uses this evidence when choosing subsequent candidates.
In \method, a deterministic mapping $\phi$ converts each log into a log profile, giving the history
\[
  \mathcal{D}^{\mathrm{log}}_k=\{(c_i,\,t_i,\,\phi(L_i))\}_{i=1}^k,
\]
where $L_i=L(c_i;P)$.
The log profile supplies evidence for a hypothesis about solver behaviour and a prediction that a subsequent evaluation can confirm or refute.
This changes the information available to the tuner while retaining the same tuning objective.
\Cref{tab:observes} compares the information and outputs of different tuners.

\begin{table}[t]
\centering
\small
\setlength{\tabcolsep}{3.5pt}
\begin{tabular}{@{}lccccc@{}}
\toprule
\textbf{Tuner} & \textbf{Auto.} & \textbf{Know.} & \textbf{Inst.} & \textbf{Log} & \textbf{Hyp.} \\
\midrule
BO (SMAC) & \cmark & \xmark & \xmark & \xmark & \xmark \\
LLM-BO (\grimip) & \cmark & \cmark & \cmark & \xmark & \xmark \\
Human expert & \xmark & \cmark & \cmark & \cmark & \cmark \\
\rowcolor{ours_color!15} \method (ours) & \cmark & \cmark & \cmark & \cmark & \cmark \\
\bottomrule
\end{tabular}
\caption{Comparison of tuners.
\textbf{Auto.}: runs without a human in the loop.
\textbf{Know.}: uses solver knowledge, such as parameter semantics.
\textbf{Inst.}: uses information about the instance before tuning (\grimip: static features and features of a 30\,s default run; \method: structure profile and log profile of the default run).
\textbf{Log}: uses the log of every evaluation when choosing the next candidate.
\textbf{Hyp.}: states for each candidate a hypothesis that the next evaluation can confirm or refute.}
\label{tab:observes}
\end{table}

%% file: sections/method.tex
\section{Methodology}
\label{sec:method}

\begin{figure*}[t]
  \centering
  \includegraphics[width=\textwidth]{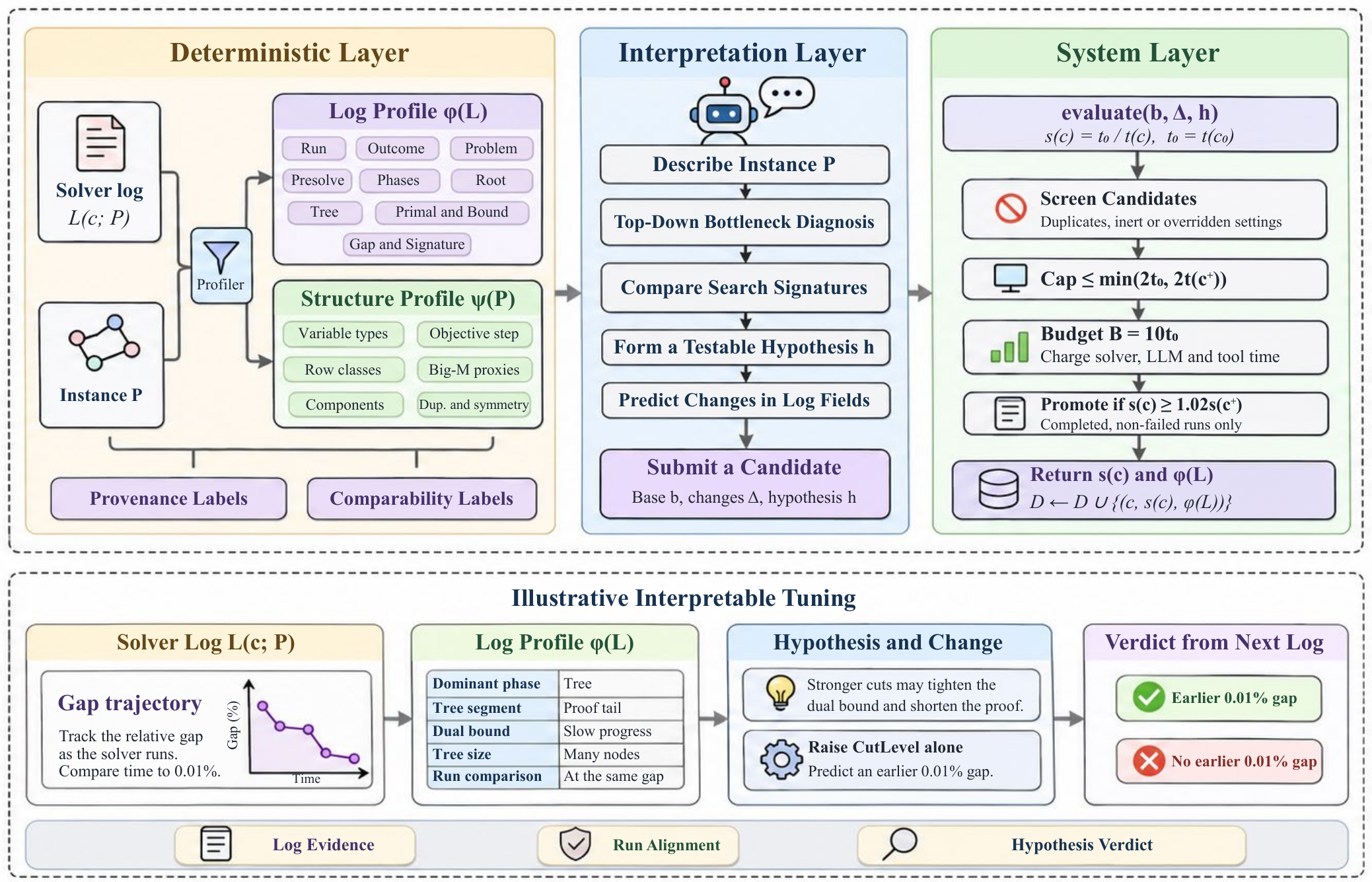}
  \caption{Overview of \method.
  Top: each evaluation passes through the deterministic layer (\cref{sec:method:profile}), the interpretation layer (\cref{sec:method:interp}), and the system layer (\cref{sec:method:system}), which returns $s(c)$ and $\phi(L)$ to the agent.
  Bottom: an illustrative tuning step; the log fields and outcomes are schematic, not experimental results.}
  \label{fig:overview}
\end{figure*}

\method follows the practice of GPU kernel optimisation: a profiler turns raw counters into metrics with a known meaning.
\method splits the work into three layers accordingly.
The deterministic layer plays the role of the profiler, the interpretation layer teaches the agent how to read its output, and the system layer keeps the decisions that should not depend on the agent at all; \cref{fig:overview} gives an overview and \cref{alg:loop} summarises the loop.

\subsection{Deterministic Layer}
\label{sec:method:profile}

\paragraph{Why deterministic code, not the agent.}
Reading error arises while the agent computes a quantity from progress rows.
Code can compute each quantity the same way on every log, defines conventions such as where the root ends once, spends no LLM time that would count against the budget, and makes fields comparable across evaluations and instances.
The agent should read the profiles and open the log only for details.

\paragraph{Log profile.}
After every evaluation, the host parses the log into a log profile $\phi(L)$ (\cref{tab:profile_groups}).
It also computes a \emph{base comparison} against the log of the base configuration, locating differences in the printed search path and solving time and comparing truncated runs at the same gap.

\paragraph{Structure profile.}
Once per instance, the host computes a structure profile $\psi(P)$ of properties that do not depend on the configuration.
These properties frame hypotheses about bottlenecks and help rule out directions in the search space.

\paragraph{Provenance and comparability labels.}
Each log-profile field carries a \emph{provenance label} stating how its value was obtained and a \emph{comparability label} stating when it can be compared across runs.
The labels make the limits of the evidence explicit, so that the agent can use truncated runs without treating them as completed runs.

\begin{table}[t]
\centering
\small
\setlength{\tabcolsep}{3pt}
\begin{tabularx}{\columnwidth}{@{}lrX@{}}
\toprule
\textbf{Group} & \textbf{\#} & \textbf{Question answered} \\
\midrule
Run & 18 & Can this log be trusted and compared? \\
Outcome & 8 & How did the run end? \\
Problem & 14 & What is the instance (same for all configurations)? \\
Presolve & 8 & How much did presolve remove? \\
Phases & 14 & Where did the time go (proxies)? \\
Root & 8 & How far did the root bound move? \\
Primal & 29 & When were incumbents found; how long was the dual segment? \\
Bound & 8 & How did the dual bound move? \\
Tree & 11 & How large and costly was the tree? \\
Gap progress & 6 & When were 10/1/0.1/0.01\% gaps reached? \\
Signature & 2 & Did the printed search path change? \\
\bottomrule
\end{tabularx}
\caption{Groups of the log profile (126 fields). All fields with their labels are listed in \cref{app:fields}.}
\label{tab:profile_groups}
\end{table}

\subsection{Interpretation Layer}
\label{sec:method:interp}

\paragraph{Top-down diagnosis.}
Before each candidate, the skill guides the agent to read the profiles, check that the run is usable, diagnose the observed cost, and form a hypothesis (\cref{app:prompts:skill}).
The top-down diagnosis in \cref{tab:diagnosis_tree} proceeds from solver phases to activity within a segment and, for the tree, node volume or node cost.
Phase times are proxies and truncated runs give one-sided values, so the descent stops at the deepest level supported by the provenance and comparability labels.
Truncated runs are compared at equal gap thresholds.

A bottleneck describes observed cost without restricting the parameter search.
The agent chooses a \emph{direct} hypothesis within the observed phase, an \emph{upstream} hypothesis in an earlier phase, or an \emph{interaction} hypothesis across interacting parameters.
Before attribution, the base comparison checks the printed search path: with an identical path, a time difference spread across the run counts as noise, while one concentrated in an interval points to a change the log does not print (\cref{app:unexplained}).

\begin{table}[t]
\centering
\footnotesize
\setlength{\tabcolsep}{2pt}
\renewcommand{\arraystretch}{1.05}
\begin{tabularx}{\columnwidth}{@{}l>{\raggedright\arraybackslash}X@{}}
\toprule
\textbf{Category} & \textbf{Evidence read at the leaf} \\
\midrule
\multicolumn{2}{@{}l}{\textit{Level 1: Where the time goes}} \\
\hspace{0.6em}Presolve-bound & Time spent in presolve \\
\hspace{0.6em}Root-bound & Time spent at the root: root LP, root rounds, and the wait before the first tree row \\
\hspace{0.6em}Tree-bound & Time from the first tree row to the end \\
\midrule
\multicolumn{2}{@{}l}{\textit{Level 2: What happens in the segment}} \\
\hspace{0.6em}Presolve & Reductions achieved relative to presolve time \\
\hspace{0.6em}Root & Which root phase holds most of the root time \\
\hspace{0.6em}Tree & How the tree time splits at the last incumbent update \\
\hspace{1.2em}Primal segment & When the first incumbent appears and how fast the primal gap closes \\
\hspace{1.2em}Dual segment & How the bound climbs and how the tree ends \\
\midrule
\multicolumn{2}{@{}l}{\textit{Level 3: Node volume or node cost (tree only)}} \\
\hspace{0.6em}Node volume & Number of nodes, and the share processed after the last incumbent \\
\hspace{0.6em}Node cost & Time and LP iterations per node \\
\bottomrule
\end{tabularx}
\caption{Top-down diagnosis. Level~1 partitions the solving time; Levels~2 and~3 refine one category at a time; leaves are bottlenecks. The descent is preceded by a check that the run is usable and followed by parameter hypotheses.}
\label{tab:diagnosis_tree}
\end{table}

The agent then submits a candidate through \texttt{evaluate(base, changes, rationale)}, where the rationale states a hypothesis linking the change to the observed cost and predicts how named fields will change.
After the evaluation, the agent records a verdict on its hypothesis: confirmed, refuted or unexplained. 

\subsection{System Layer}
\label{sec:method:system}
The system layer manages the budget, promotion, candidate checks, and the agent's environment independently of the agent (\cref{alg:loop}).

\paragraph{Budget.}
The default run establishes $B=10\,t(c_0)$ and is not charged to it.
Subsequent solver, LLM, and tool time count toward the budget, and a cap limits each evaluation.

\paragraph{Promotion.}
The system uses evaluation outcomes to promote candidates and make the final selection, with speedup $s(c)=t(c_0)/t(c)$ and runs reaching the time limit assigned that limit.
Truncated runs provide partial log evidence but are ineligible for promotion and final selection (\cref{app:selection}).

\paragraph{Candidate checks.}
Before evaluation, the system rejects repeated candidates and candidates with dead or overridden parameters, and returns the reason.

\paragraph{Environment.}
The agent accesses profiles and de-identified logs through a sandbox that hides the instance name and solver licence (\cref{app:dsh}).

\begin{algorithm}[t]
\caption{\method on one instance $P$}
\label{alg:loop}
\small
\begin{algorithmic}[1]
\Require instance $P$, default configuration $c_0$, budget factor $10$
\State run $c_0$; $t_0\gets t(c_0)$; $B\gets 10\,t_0$
\State $\psi\gets$ structure profile of $P$; $\phi_0\gets$ log profile of the default run
\State $c^+\gets c_0$; $\mathcal{D}\gets\{(c_0,1,\phi_0)\}$
\While{the remaining budget allows an evaluation}
  \State agent: diagnosis on $\psi,\mathcal{D}$; submit $(b,\Delta,h)$ \Comment{base configuration, change, hypothesis}
  \If{$b\oplus\Delta$ repeats a configuration or sets dead or overridden parameters}
    \State rejection with reason; \textbf{continue}
  \EndIf
  \State run $c=b\oplus\Delta$ with cap $\min(2t_0,2t(c^+))$
  \State $\phi\gets$ log profile; $\mathcal{D}\gets\mathcal{D}\cup\{(c,s(c),\phi)\}$
  \If{the run completed and improves $c^+$ under the promotion rule} $c^+\gets c$ \Comment{promotion} \EndIf
  \State return $s(c)$ and $\phi$; agent records a verdict on $h$
\EndWhile
\State \Return $\hat c$ under the final selection rule (\cref{app:selection})
\end{algorithmic}
\end{algorithm}

%% file: sections/experiments.tex
\section{Experiments}
\label{sec:exp}

\subsection{Datasets}
\label{sec:datasets}

\begin{table}[t]
\centering
\footnotesize
\setlength{\tabcolsep}{2pt}
\renewcommand{\arraystretch}{1.12}
\begin{tabular*}{\columnwidth}{@{}l@{\extracolsep{\fill}}ccc@{}}
\toprule
\textbf{Dataset} & \textbf{Type} & \textbf{Vars} & \textbf{Cons} \\
\midrule
CORAL\textsuperscript{a} & Medium & [62, 573K] & [6, 550K] \\
MIRP\textsuperscript{b} & Hard & [2K, 194K] & [1K, 109K] \\
MIPLIB~2017\textsuperscript{c} & Hard & [30, 1.43M] & [4, 2.90M] \\
\midrule
Item Placement\textsuperscript{d} & Hard & 1,083 & 195 \\
Load Balancing\textsuperscript{d} & Hard & 61,000 & 64,304 \\
Anonymous\textsuperscript{d} & Hard & 37,881 & 49,603 \\
\bottomrule
\end{tabular*}
\par\raggedright\scriptsize
\textsuperscript{a}\citep{gomes2008connections}; mean Vars $=19{,}989$, Cons $=12{,}482$.
\textsuperscript{b}\citep{papageorgiou2014mirplib}; mean Vars $=34{,}193$, Cons $=16{,}170$.
\textsuperscript{c}\citep{MIPLIB2017}, benchmark set; mean Vars $=51{,}422$, Cons $=56{,}876$.
\textsuperscript{d}\citep{gasse2022ml4co}.
\caption{Summary of datasets.}
\label{tab:dataset_summary}
\end{table}

Our evaluation draws on six datasets (\cref{tab:dataset_summary}): a medium dataset (CORAL), a structured hard dataset (MIRP), three ML4CO datasets (Item Placement, Load Balancing, and Anonymous), and, as a heterogeneous hard dataset, the full MIPLIB~2017 benchmark set of 240 instances~\citep{MIPLIB2017}.
The first five follow the benchmark of \citet{luo2026grimip}, which uses a MIPLIB subset instead.
Sample coverage is detailed in \cref{app:dataset_counts}.

\subsection{Baselines}
\label{sec:baselines}
We compare \method with the following baselines and reference:
\textcircled{\scriptsize 1}~\textbf{Default}: the default configuration of COPT.
\textcircled{\scriptsize 2}~\textbf{SMAC}~\citep{lindauer2022smac3}: SMAC applied to each instance separately, over the six main COPT parameters.
\textcircled{\scriptsize 3}~\mbox{\textbf{Random}}: random search over the same six-dimensional search space as SMAC, with the same parameter domains.
\textcircled{\scriptsize 4}~\textbf{\grimip}~\citep{luo2026grimip}: LLM-driven BO with automated search-space selection and warm-starting, ported to COPT.
It selects its search space from the 42 COPT parameters of \method (\cref{app:space}).
\textcircled{\scriptsize 5}~\textbf{Best Observed}: a retrospective reference giving the best eligible outcome for each instance across the configurations evaluated by all compared methods, including Default (\cref{app:metrics}).

All tuners share the run settings and budget and the final selection; both LLM-based tuners use the same LLM, DeepSeek-V4.1-Flash.
Details of each baseline are given in \cref{app:baselines}.

\subsection{Setup}
\label{sec:exp:setup}
Each run uses 8 threads bound to a fixed group of 8 cores and a time limit of 1800\,s.
Hardware and software are listed in \cref{app:env}.
Each instance receives a budget of $B=10\,t(c_0)$.

\subsection{Main Results}

\begin{table*}[t]
\centering
\footnotesize
\newcommand{\apecresult}[2]{\resultshade{#1}{{\fontsize{9.6}{11}\selectfont#2}}}
\setlength{\tabcolsep}{0.5pt}
\begin{tabularx}{\textwidth}{@{}l *{14}{C}@{}}
\toprule
& \multicolumn{7}{c}{\textbf{Item Placement}~\citep{gasse2022ml4co}} & \multicolumn{7}{c}{\textbf{Load Balancing}~\citep{gasse2022ml4co}} \\
\cmidrule(lr){2-8}\cmidrule(lr){9-15}
\textbf{Method} & Time $\downarrow$ & Imp. $\uparrow$ & Eff. $\uparrow$ & Solv. $\uparrow$ & Gap $\downarrow$ & Par. $\downarrow$ & Step $\downarrow$ & Time $\downarrow$ & Imp. $\uparrow$ & Eff. $\uparrow$ & Solv. $\uparrow$ & Gap $\downarrow$ & Par. $\downarrow$ & Step $\downarrow$ \\
\midrule
Default & \resultshade{0}{1800.00} & \resultshade{0}{1.00$\times$} & -- & \resultshade{0}{0.0} & \resultshade{0}{35.29} & -- & -- & \resultshade{0}{1616.91} & \resultshade{0}{1.00$\times$} & -- & \resultshade{0}{29.8} & \resultshade{0}{0.32} & -- & -- \\
B. Obs. & 1776.50 & 1.02$\times$ & 90.0 & 2.0 & 17.24 & -- & -- & 1437.70 & 1.17$\times$ & 97.9 & 44.7 & 0.22 & -- & -- \\
\cmidrule(r){1-1}\cmidrule(lr){2-15}
Random & \resultshade{100}{1776.50} & \resultshade{100}{1.02$\times$} & \resultshade{67}{44.0} & \resultshade{100}{2.0} & \resultshade{75}{28.16} & \resultshade{67}{4.64} & \resultshade{0}{4.00} & \resultshade{25}{1572.77} & \resultshade{33}{1.03$\times$} & \resultshade{0}{23.4} & \resultshade{33}{34.0} & \resultshade{33}{0.30} & \resultshade{0}{5.64} & \resultshade{0}{4.14} \\
SMAC & \resultshade{0}{1800.00} & \resultshade{0}{1.00$\times$} & \resultshade{0}{28.0} & \resultshade{0}{0.0} & \resultshade{25}{30.35} & \resultshade{100}{4.57} & \resultshade{33}{3.90} & \resultshade{75}{1532.36} & \resultshade{67}{1.07$\times$} & \resultshade{50}{70.2} & \resultshade{67}{38.3} & \resultshade{67}{0.26} & \resultshade{67}{4.76} & \resultshade{33}{3.98} \\
\grimip & \resultshade{0}{1800.00} & \resultshade{0}{1.00$\times$} & \resultshade{33}{32.0} & \resultshade{0}{0.0} & \resultshade{50}{28.26} & \resultshade{0}{5.88} & \resultshade{67}{3.58} & \resultshade{50}{1543.24} & \resultshade{67}{1.07$\times$} & \resultshade{50}{70.2} & \resultshade{67}{38.3} & \resultshade{67}{0.26} & \resultshade{33}{5.48} & \resultshade{67}{3.39} \\
{\fontsize{9.6}{11}\selectfont\method} & \apecresult{50}{1791.09} & \apecresult{50}{1.01$\times$} & \apecresult{100}{\textbf{90.0}} & \apecresult{100}{2.0} & \apecresult{100}{\textbf{18.67}} & \apecresult{33}{5.07} & \apecresult{100}{\textbf{2.17}} & \apecresult{100}{1456.80} & \apecresult{100}{\textbf{1.15}$\times$} & \apecresult{100}{95.7} & \apecresult{100}{\textbf{44.7}} & \apecresult{100}{0.22} & \apecresult{100}{4.31} & \apecresult{100}{\textbf{1.86}} \\
\bottomrule
\end{tabularx}

\vspace{4pt}
\begin{tabularx}{\textwidth}{@{}l *{14}{C}@{}}
\toprule
& \multicolumn{7}{c}{\textbf{Anonymous}~\citep{gasse2022ml4co}} & \multicolumn{7}{c}{\textbf{CORAL}~\citep{gomes2008connections}} \\
\cmidrule(lr){2-8}\cmidrule(lr){9-15}
\textbf{Method} & Time $\downarrow$ & Imp. $\uparrow$ & Eff. $\uparrow$ & Solv. $\uparrow$ & Gap $\downarrow$ & Par. $\downarrow$ & Step $\downarrow$ & Time $\downarrow$ & Imp. $\uparrow$ & Eff. $\uparrow$ & Solv. $\uparrow$ & Gap $\downarrow$ & Par. $\downarrow$ & Step $\downarrow$ \\
\midrule
Default & \resultshade{0}{1047.77} & \resultshade{0}{1.00$\times$} & -- & \resultshade{0}{50.0} & \resultshade{0}{21.66} & -- & -- & \resultshade{0}{414.33} & \resultshade{0}{1.00$\times$} & -- & \resultshade{0}{80.0} & \resultshade{0}{5.82} & -- & -- \\
B. Obs. & 716.54 & 2.37$\times$ & 100.0 & 66.0 & 5.62 & -- & -- & 315.49 & 1.56$\times$ & 93.0 & 85.0 & 3.82 & -- & -- \\
\cmidrule(r){1-1}\cmidrule(lr){2-15}
Random & \resultshade{25}{877.93} & \resultshade{25}{1.38$\times$} & \resultshade{0}{44.0} & \resultshade{33}{62.0} & \resultshade{25}{12.82} & \resultshade{67}{5.03} & \resultshade{33}{3.95} & \resultshade{25}{373.45} & \resultshade{25}{1.09$\times$} & \resultshade{0}{45.0} & \resultshade{100}{84.0} & \resultshade{33}{4.97} & \resultshade{33}{5.35} & \resultshade{0}{4.34} \\
SMAC & \resultshade{100}{762.95} & \resultshade{75}{1.92$\times$} & \resultshade{67}{94.0} & \resultshade{67}{64.0} & \resultshade{50}{7.55} & \resultshade{33}{5.19} & \resultshade{0}{4.03} & \resultshade{50}{351.32} & \resultshade{50}{1.27$\times$} & \resultshade{100}{75.0} & \resultshade{50}{83.0} & \resultshade{100}{4.07} & \resultshade{67}{5.33} & \resultshade{33}{4.27} \\
\grimip & \resultshade{50}{772.93} & \resultshade{50}{1.91$\times$} & \resultshade{33}{88.0} & \resultshade{100}{66.0} & \resultshade{75}{6.80} & \resultshade{0}{5.98} & \resultshade{67}{3.32} & \resultshade{75}{345.52} & \resultshade{75}{1.30$\times$} & \resultshade{33}{54.0} & \resultshade{50}{83.0} & \resultshade{67}{4.08} & \resultshade{0}{5.90} & \resultshade{67}{3.97} \\
{\fontsize{9.6}{11}\selectfont\method} & \apecresult{75}{772.71} & \apecresult{100}{1.96$\times$} & \apecresult{100}{98.0} & \apecresult{100}{66.0} & \apecresult{100}{6.56} & \apecresult{100}{4.90} & \apecresult{100}{\textbf{1.74}} & \apecresult{100}{339.92} & \apecresult{100}{\textbf{1.40}$\times$} & \apecresult{67}{70.0} & \apecresult{50}{83.0} & \apecresult{100}{4.07} & \apecresult{100}{\textbf{3.23}} & \apecresult{100}{\textbf{1.94}} \\
\bottomrule
\end{tabularx}

\vspace{4pt}
\begin{tabularx}{\textwidth}{@{}l *{14}{C}@{}}
\toprule
& \multicolumn{7}{c}{\textbf{MIRP}~\citep{papageorgiou2014mirplib}} & \multicolumn{7}{c}{\textbf{MIPLIB~2017}~\citep{MIPLIB2017}} \\
\cmidrule(lr){2-8}\cmidrule(lr){9-15}
\textbf{Method} & Time $\downarrow$ & Imp. $\uparrow$ & Eff. $\uparrow$ & Solv. $\uparrow$ & Gap $\downarrow$ & Par. $\downarrow$ & Step $\downarrow$ & Time $\downarrow$ & Imp. $\uparrow$ & Eff. $\uparrow$ & Solv. $\uparrow$ & Gap $\downarrow$ & Par. $\downarrow$ & Step $\downarrow$ \\
\midrule
Default & \resultshade{0}{1782.38} & \resultshade{0}{1.00$\times$} & -- & \resultshade{50}{2.0} & \resultshade{0}{48.56} & -- & -- & \resultshade{50}{410.28} & \resultshade{0}{1.00$\times$} & -- & \resultshade{67}{85.6} & \resultshade{0}{9.93} & -- & -- \\
B. Obs. & 1769.41 & 1.02$\times$ & 72.0 & 2.0 & 27.71 & -- & -- & 319.51 & 1.78$\times$ & 92.8 & 88.1 & 5.02 & -- & -- \\
\cmidrule(r){1-1}\cmidrule(lr){2-15}
Random & \resultshade{25}{1780.22} & \resultshade{0}{1.00$\times$} & \resultshade{0}{20.0} & \resultshade{50}{2.0} & \resultshade{25}{46.46} & \resultshade{100}{5.20} & \resultshade{0}{4.00} & \resultshade{0}{425.11} & \resultshade{25}{1.06$\times$} & \resultshade{0}{36.9} & \resultshade{0}{84.3} & \resultshade{25}{8.66} & \resultshade{33}{5.28} & \resultshade{0}{4.29} \\
SMAC & \resultshade{75}{1770.70} & \resultshade{100}{1.02$\times$} & \resultshade{33}{64.0} & \resultshade{50}{2.0} & \resultshade{100}{31.64} & \resultshade{67}{5.25} & \resultshade{33}{3.90} & \resultshade{25}{412.83} & \resultshade{50}{1.15$\times$} & \resultshade{33}{50.0} & \resultshade{33}{84.7} & \resultshade{50}{8.02} & \resultshade{67}{5.08} & \resultshade{33}{4.20} \\
\grimip & \resultshade{50}{1772.14} & \resultshade{100}{1.02$\times$} & \resultshade{67}{66.0} & \resultshade{50}{2.0} & \resultshade{50}{34.52} & \resultshade{0}{5.94} & \resultshade{67}{3.59} & \resultshade{75}{399.16} & \resultshade{75}{1.16$\times$} & \resultshade{67}{56.4} & \resultshade{67}{85.6} & \resultshade{75}{7.38} & \resultshade{0}{5.74} & \resultshade{67}{3.83} \\
{\fontsize{9.6}{11}\selectfont\method} & \apecresult{100}{1769.41} & \apecresult{100}{1.02$\times$} & \apecresult{100}{68.0} & \apecresult{50}{2.0} & \apecresult{75}{32.72} & \apecresult{33}{5.29} & \apecresult{100}{\textbf{2.26}} & \apecresult{100}{332.70} & \apecresult{100}{\textbf{1.58}$\times$} & \apecresult{100}{84.3} & \apecresult{100}{88.1} & \apecresult{100}{\textbf{5.16}} & \apecresult{100}{\textbf{3.15}} & \apecresult{100}{\textbf{1.57}} \\
\bottomrule
\end{tabularx}
\caption{Comparison of tuners across datasets. \textbf{Time}: mean solving time (s), with timeouts counted as 1800\,s; \textbf{Imp.}: SGM10 of Default solving times divided by SGM10 of selected-configuration solving times (shift 10\,s); \textbf{Eff.}: instances on which the selected configuration improves on Default (\%); \textbf{Solv.}: instances solved within the time limit (\%); \textbf{Gap}: shifted geometric mean (shift 1 percentage point) of the final relative gap (\%) on instances where Default times out, with gaps above 100\% and runs without an incumbent counted as 100\%; \textbf{Par.}: mean non-default parameters in the selected configuration, over instances with a non-default selection only; \textbf{Step}: parameters changed per evaluation, measured against the nearest previously evaluated configuration and averaged within each instance, then over instances with evaluations (\cref{sec:interp:measures}). Darker backgrounds indicate better ranks within each dataset and metric; ties share a color. \textbf{B. Obs.}: Best Observed, excluded from ranking; \textbf{Par.}\ and \textbf{Step} are omitted for this retrospective reference. Sample counts and calculation details are given in \cref{app:dataset_counts,app:metrics}.}
\label{tab:main}
\end{table*}

\paragraph{Overall performance.}
\Cref{tab:main} shows that \method improves over Default on all six datasets and has the highest Imp.\ among tuners on Load Balancing, Anonymous, CORAL, and MIPLIB~2017.
On MIPLIB~2017, it reaches 1.58$\times$, compared with 1.15$\times$ for SMAC and 1.16$\times$ for \grimip.
Random achieves the highest Imp.\ on Item Placement, while \method, SMAC, and \grimip have similar Imp.\ on MIRP.

\paragraph{Readable configurations and searches.}
\method averages fewer step changes than SMAC and \grimip on all six datasets and fewer final changes among non-default selections on four, even though it searches 42 parameters, compared with six for SMAC.
These counts measure readability; \cref{app:case} examines the reasoning behind selected changes.

\begin{figure}[t]
  \centering
  \includegraphics[width=\columnwidth]{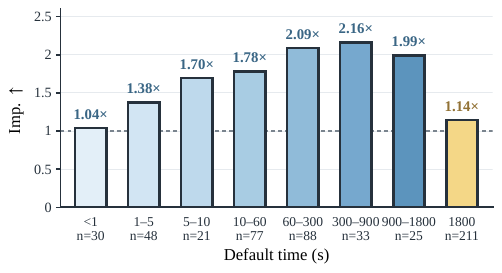}
  \caption{Performance of \method by default time, pooling 533 instances across the six datasets. Imp.\ is recomputed within each group as in \cref{tab:main}, with equal weight per instance; $n$ denotes the group size. Intervals include the lower endpoint and exclude the upper endpoint. Blue darkens with increasing default time; yellow marks default runs reaching the 1800\,s time limit.}
  \label{fig:default-time}
\end{figure}

\paragraph{Performance and instance difficulty.}
\Cref{fig:default-time} groups instances by default time, a measure of their difficulty for the default configuration.
Imp.\ is 1.04$\times$ below 1\,s and reaches 2.16$\times$ in the 300--900\,s group.
For instances whose solving times are far beyond the 1800\,s limit, even a large reduction may leave a run above the limit.
In the default time-limit group, \method solves 25 of 211 instances; the remaining 186 are counted at 1800\,s under both configurations, contributing no solving-time improvement to Imp.
Imp.\ is therefore 1.14$\times$, while \method reduces Gap from 14.45\% to 8.40\%, a relative reduction of 41.9\%.
The groups differ in both dataset composition and budget, which scales with default time; these comparisons describe the observed results without isolating the effect of either factor.

\subsection{Performance Across the Budget}
\label{sec:budget-curves}

\begin{figure}[t]
  \centering
  \includegraphics[width=\columnwidth]{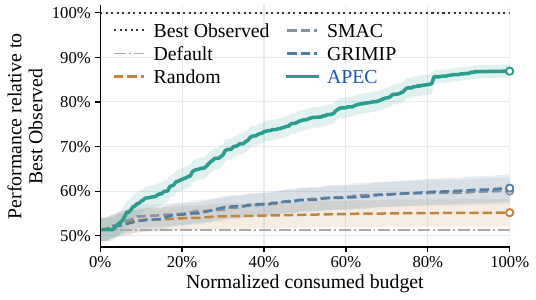}
  \caption{Performance across the budget on 209 MIPLIB~2017 instances, relative to Best Observed (100\%). The solid line shows \method; dashed curves show the baselines. Circles mark final selections. Shading shows $\pm1$ bootstrap standard deviation. Coverage and metric definitions are in \cref{app:budget-curves}.}
  \label{fig:budget-curves}
\end{figure}

\Cref{fig:budget-curves} follows performance relative to Best Observed over the normalized consumed budget, including solver, LLM, and tool time.
\method reaches 76.0\% of Best Observed performance at half of the budget and 86.9\% at the full budget on this common subset.
All methods use the same 209 instances.

\subsection{Ablation Studies}
\label{sec:ablation}

\begin{table}[t]
\centering
\footnotesize
\setlength{\tabcolsep}{2pt}
\begin{tabularx}{\columnwidth}{@{}>{\raggedright\arraybackslash}X*{4}{r}@{}}
\toprule
\textbf{Variant} & Imp. $\uparrow$ & Eff. $\uparrow$ & Par. $\downarrow$ & Step $\downarrow$ \\
\midrule
Default & 1.00$\times$ & -- & -- & -- \\
\method & 1.80$\times$ & 84.5 & 2.96 & 1.51 \\
\quad raw-log variant & 1.62$\times$ & 75.4 & 3.87 & 1.98 \\
\quad w/o interpretation layer & 1.68$\times$ & 80.7 & 3.63 & 1.77 \\
\bottomrule
\end{tabularx}
\caption{Ablations on 187 instances from the MIPLIB~2017 benchmark set with default time at most 600\,s. Setup and metrics follow \cref{sec:exp:setup} and \cref{tab:main}, respectively.}
\label{tab:ablation}
\end{table}

The \emph{raw-log variant} provides de-identified raw logs and shell access.
The \emph{w/o interpretation layer} variant retains the log profiles, structure profiles, and base comparisons, and replaces the skill for diagnosis and hypothesis checking (\cref{sec:method:interp}) with field descriptions (\cref{app:fields}).

Default and all three tuners solve every instance in this subset, so the performance differences in \cref{tab:ablation} reflect changes in solving time on already-solvable instances.
In the paired comparison, \method selects faster configurations more often than either variant, with a narrower margin against w/o interpretation layer than against the raw-log variant (\cref{app:dataset_counts}).

The solving-time advantage accompanies fewer step changes and fewer final changes among non-default selections.
When evaluated configurations differ in fewer parameters, there are fewer simultaneous changes to consider when interpreting differences in their logs.
The interpretation layer gives the agent a way to use such comparisons by linking a controlled parameter change to a prediction for a named field in the next log profile.

%% file: sections/interpretability.tex
\section{Towards Interpretable Tuning}
\label{sec:interp}

A black-box tuner can be judged only by its selected configuration.
A log-grounded agent can also be judged by its reasoning, because each evaluation carries a hypothesis that the next log can refute.

\subsection{Measuring Interpretability}
\label{sec:interp:measures}
We use final changes and step changes to assess the readability of a tuner's output and search, respectively.
As shown in \cref{tab:main}, \method has fewer mean step changes than SMAC and \grimip on all six datasets, and fewer mean final changes on four.

\subsection{Inspecting the Reasoning}
\label{sec:interp:reasoning}
The counts describe readability, but do not establish that an explanation is correct.

\begin{figure}[t]
  \centering
  \includegraphics[width=\columnwidth]{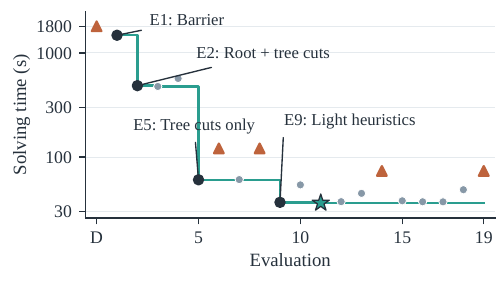}
  \caption{Tuning trajectory on \texttt{anonymous\_33}. D denotes Default. Triangles mark run limits (lower bounds), the line traces the best completed solving time, and the star marks the selected evaluation (E11, 36.3\,s). The vertical axis is logarithmic.}
  \label{fig:anonymous-trajectory}
\end{figure}

On \texttt{anonymous\_33}, the default log showed that the tree dominated solving time, with 57 LP iterations per node (\cref{fig:anonymous-trajectory}).
\method tried barrier to lower node cost, which fell from 2.47 to 1.44\,ms (E1).
The run still processed over a million nodes, so the agent tried stronger cuts (E2) and then tested root and tree cuts separately.
Keeping only tree cuts reduced solving time from 484.5\,s with both to 60.5\,s (E5).
The E5 log showed that most nodes were processed after the last incumbent update, so the agent tried stronger heuristics to find a better incumbent earlier.
Node cost rose to 12.0\,ms (E8), prompting a test of lighter heuristics that brought solving time down to 36.8\,s (E9).
\Cref{app:case:anonymous} gives the full comparison.

The verdict audit links recorded verdicts to their predictions. Experts review a stratified sample against solver logs and run comparisons (\cref{app:verdict-audit}).
The audit judged 45 of the 60 sampled explanations supported, nine contradicted, and six insufficient.
For example, one verdict attributed a faster run to cheaper nodes even though node cost increased.

%% file: sections/related_work.tex
\section{Related Work}
\label{sec:related}

\paragraph{Algorithm configuration for MIP solvers.}
ParamILS~\citep{hutter2009paramils}, irace~\begin{NoHyper}\citep{lopez-ibanez2016irace}\end{NoHyper}, and SMAC~\citep{hutter2011sequential,lindauer2022smac3} search for configurations using performance observations, with automated tuning also studied specifically for MIP solvers~\citep{hutter2010automated}.
Instance-specific approaches condition configuration choices on instance properties through clustering~\citep{kadioglu2010isac} or learned graph representations~\citep{valentin2022instancewise}.
These approaches address which configuration performs well; our focus is on using solver logs to make the tuning process interpretable.

\paragraph{Run-time statistics as features.}
Algorithm selection and runtime prediction compute \emph{probing features} by running a solver briefly and recording statistics of the run~\citep{xu2008satzilla,hutter2014algorithm}.
Such features provide inputs to predictive models, establishing that solver behaviour can inform algorithm selection and performance prediction.
\method uses the logs produced by candidate evaluations as evidence for diagnosis, with provenance and comparability labels that specify how individual fields can support comparisons between runs.

\paragraph{LLMs for tuning.}
LLMs recommend configurations directly~\citep{zhang2023using,kochnev2025optuna} or support BO through initialization, candidate sampling, and performance prediction~\citep{liu2024llmbo,mahammadli2024sequential}.
The closest work, \grimip~\citep{luo2026grimip}, uses an LLM throughout a BO loop for MIP solvers.
It extracts features from a short default run to select the search space and initialise the search, then observes scalar performance feedback from subsequent evaluations.
\method instead uses each evaluation's log to formulate and check a hypothesis about solver behaviour, exposing the evidence behind parameter choices.
Concurrently, \citet{you2026ldst} also use an LLM to inspect solver logs and propose sparse parameter changes for MIP solvers.

Complementary work is discussed in \cref{app:related}.

%% file: sections/conclusion.tex
\section{Conclusion}
\label{sec:conclusion}

We present \method, a log-grounded agent for instance-specific MIP
solver tuning. Inspired by expert profiling, \method combines log and
structure profiles with hypothesis-driven diagnosis and controlled
evaluations, bringing evidence of solver behavior into the tuning loop
and making parameter changes available for inspection.
Across six datasets, \method improves over Default on all datasets and
combines strong performance with more readable configurations and tuning
searches. 
However, solver behavior remains subject to
uncertainty, and some outcomes may not be explainable from the available
logs. A promising direction for future work may be equipping \method with a memory
mechanism that uses experience from similar instances to recommend
parameter. 

%% file: sections/limitations.tex
\section*{Limitations}

The default COPT log omits statistics for some mechanisms, limiting diagnosis and causal attribution.
Fewer parameter changes do not prove explanation correctness, and the verdict audit covers a limited recording channel and a stratified sample reviewed by experts (\cref{app:verdict-audit}).
Our experiments use one solver and one LLM.
Each candidate is evaluated once, so performance variability can affect selection; the agent also incurs LLM usage cost.

%% file: sections/appendix.tex
\section{Experiment Environment}
\label{app:env}
All experiments run on a server with Ubuntu 24.04.3 LTS, an AMD EPYC 9754 processor (128 cores, 256 hardware threads, 256\,MiB L3 cache), and 251\,GiB of RAM.
Each run uses 8 threads bound to a fixed group of 8 cores; concurrent tasks never share cores.
Software: COPT~8.1, Python~3.11, and SMAC3~2.3.1. LLM: DeepSeek-V4.1-Flash.

\subsection{Evaluation and Selection Rules}
\label{app:selection}
The incumbent configuration $c^+$ is initialized to $c_0$.
Each evaluation is subject to a cap of $\min(2t(c_0),\,2t(c^+))$, in addition to the time limit and the remaining budget.
A run stopped by the cap or by budget exhaustion is a truncated run: its observed duration is only a lower bound on $t(c)$, so it is ineligible for promotion and final selection, although its log remains available to the agent as partial evidence.
The system layer rejects a candidate before running it if the candidate repeats an evaluated configuration or sets dead or overridden parameters, and states the reason.

When the default run is solved, a completed, non-failed evaluation promotes its candidate $c$ when $s(c)\ge1.02\,s(c^+)$; the margin of 2\% guards against small timing differences.
Final selection considers all completed, non-failed evaluations, including candidates that were not promoted: $\hat c$ has the highest speedup if it is at least $1.02$; otherwise, $\hat c=c_0$.

When the default run reaches the time limit, promotion and final selection instead rank solved candidates by solving time, followed by candidates reaching the full time limit ranked by final gap (missing gap is $\infty$).
Promotion requires a strict improvement over $c^+$.
Final selection includes $c_0$: an eligible candidate replaces it only if the candidate is solved or has a strictly smaller final gap; equal gaps retain $c_0$.
Runs stopped at a shorter cap or by budget exhaustion remain ineligible.
The selected configuration is not run again for confirmation.

\section{Case Study}
\label{app:case}
\begin{figure}[htbp]
\centering
\includegraphics[width=\textwidth]{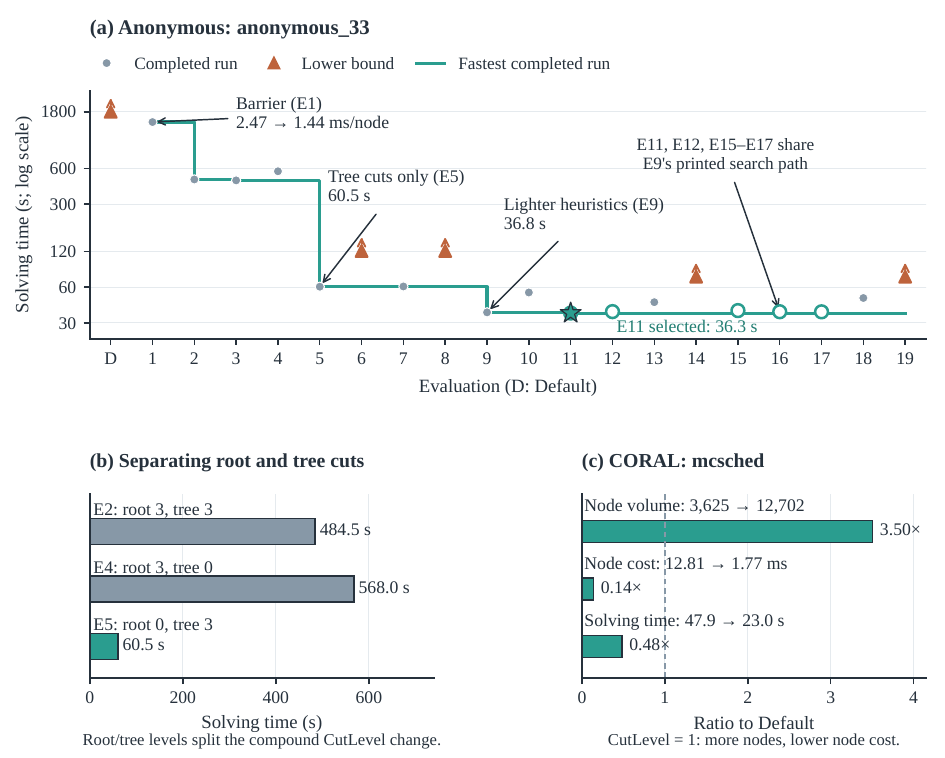}
\caption{Log-grounded tuning in two case studies. (a) All 19 evaluations on \texttt{anonymous\_33}; D denotes Default. Upward triangles mark run limits, which are lower bounds on completed solving times. The solid line follows the fastest completed run. Open circles share E9's printed search path; the star marks the selected E11. (b) Splitting the compound \texttt{CutLevel} change separates the effects of root and tree cuts. (c) On \texttt{mcsched}, setting \texttt{CutLevel}=1 increases node volume while lowering node cost and solving time. Ratios are relative to Default; node cost is the tree time per node reported in the log profile.}
\label{fig:case-study}
\end{figure}

We examine two trajectories to show how the agent's hypotheses can be checked against the subsequent logs.
\Cref{fig:case-study} visualizes the trajectory on Anonymous \texttt{anonymous\_33} and the tradeoff between node volume and node cost on CORAL \texttt{mcsched}.
These cases illustrate the recorded reasoning; \cref{app:verdict-audit} evaluates their typicality and audits verdicts beyond these examples.

\subsection{Anonymous: Separating Root and Tree Cuts}
\label{app:case:anonymous}
We follow \method on the Anonymous instance \texttt{anonymous\_33}, where its selected configuration solves the instance in 36.3\,s against 61.7\,s for \grimip.
\Cref{tab:case} pairs each change and hypothesis with the solving time, node count, and evidence in the log profile.
The default run reaches the time limit after 727,535 nodes with a final gap of 4.42\%.
Its log profile places almost all of the time in the tree, at 2.47\,ms and 57 LP iterations per node, with 92\% of the nodes processed after the last incumbent update.

\paragraph{Reading the node cost.}
The first candidate follows the diagnosis to a high node cost, and E1 confirms the prediction on the named fields: barrier lowers the node cost from 2.47 to 1.44\,ms, and the run completes within the time limit.
E2 adds aggressive cuts against the remaining node volume and reduces the solving time to 484.5\,s.

\paragraph{Splitting a compound change.}
\texttt{CutLevel} sets root and tree cuts together, so E2 does not say which half produced its gain.
The agent stated a test before running it: if root cuts drove the gain, keeping only root cuts should stay near 485\,s.
E4 keeps only root cuts and takes 568.0\,s; E5 keeps only tree cuts and takes 60.5\,s, eight times faster than E2.
The aggressive root cuts had been hiding the gain of the tree cuts, and E6 confirms that even light root cuts lead to a truncated run.
E7 then removes barrier from E5 and returns the same printed search path: the change that began the descent is no longer part of the mechanism.

\paragraph{Bracketing a level.}
Aggressive heuristics (E8) raise the node cost to 12.0\,ms; the agent read this field as heuristic overhead and tried the opposite level.
\texttt{HeurLevel}=1 (E9) lowers the node cost to 0.35\,ms and the solving time to 36.8\,s, and both neighbouring levels are worse (E10, E14).

\paragraph{Recognising no-op changes.}
Five later candidates on E9, covering restarts, branching, \texttt{Dualize}, and lighter tree cut levels, return the printed search path of E9 with identical node counts.
Before E17, the agent predicted that identical printed search paths for tree cut levels 1 to 3 would show that the tree cut level no longer matters, and the log confirmed it.
With the remaining levers ruled out, the agent stopped after 19 evaluations and 22\% of the budget; final selection returns E11, which repeats the printed search path of E9.

\paragraph{Black-box baselines.}
\grimip used all 50 evaluations and 46\% of the budget.
Its search space, chosen by automated search-space selection, contains \texttt{CutLevel} and \texttt{RootCutLevel} but not \texttt{TreeCutLevel}, and a history of configurations and solving times cannot tell which half of a cut level helps.
SMAC used 30 evaluations and nearly the whole budget and reached 44.7\,s with a configuration that switches off cuts, presolve, and strong branching together.

\begin{table}[htbp]
\centering
\footnotesize
\setlength{\tabcolsep}{3pt}
\renewcommand{\arraystretch}{1.08}
\begin{tabularx}{\textwidth}{@{}>{\raggedright\arraybackslash}p{0.08\textwidth} l >{\raggedright\arraybackslash}X >{\raggedright\arraybackslash}X r r >{\raggedright\arraybackslash}X@{}}
\toprule
\textbf{Eval.} & \textbf{Base} & \textbf{Change} & \textbf{Hypothesis in the rationale (abridged)} & \textbf{Time (s)} & \textbf{Nodes} & \textbf{What the log shows} \\
\midrule
Default & -- & -- & -- & $\ge$1800 & 727.5K & Tree-bound; 2.47\,ms and 57 LP iterations per node \\
E1 & $c_0$ & \texttt{LpMethod}=2 & Node cost is high; barrier makes nodes cheaper & 1475.9 & 1.02M & Confirmed: 1.44\,ms per node; solved within the time limit \\
E2 & E1 & \texttt{CutLevel}=3 & Node volume remains; stronger cuts prune more & 484.5 & 596.7K & Promoted; 0.81\,ms per node \\
E3 & E2 & \texttt{StrongBranching}=3, \texttt{ConflictAnalysis}=1 & Stronger branching prunes more & 475.9 & 596.7K & Printed search path equals E2's: a no-op change \\
E4 & E2 & root cuts 3, tree cuts off & Split E2: near 485\,s if root cuts drove the gain & 568.0 & 688.9K & Root cuts are not the source \\
E5 & E2 & root cuts off, tree cuts 3 & The complementary half of the split & 60.5 & 143.5K & Promoted; 8$\times$ faster than E2 \\
E6 & E5 & \texttt{RootCutLevel}=1 & Is a little root strengthening useful? & $\ge$121.0 & 281.6K & No: truncated run \\
E7 & E5 & \texttt{LpMethod} back to default & Is barrier still needed? & 60.8 & 143.5K & Printed search path equals E5's: barrier no longer matters \\
E8 & E5 & \texttt{HeurLevel}=3 & Earlier incumbents prune more & $\ge$121.1 & 9.9K & Refuted: 12.0\,ms per node \\
E9 & E5 & \texttt{HeurLevel}=1 & Lighter heuristics lower the node cost & 36.8 & 102.2K & Promoted; 0.35\,ms per node \\
E10, E14 & E9 & \texttt{HeurLevel}=0; =2 & Neighbouring heuristic levels & 54.1; $\ge$73.6 & -- & Both worse: level 1 is best \\
E11, E12, E15--E17 & E9 & restart, branching, \texttt{Dualize}, tree cut level 2 or 1 & Remaining node-cost or node-volume levers & 36.3--38.1 & 102.2K & Printed search path equals E9's: five no-op changes \\
\bottomrule
\end{tabularx}
\caption{Excerpt of the trajectory of \method on Anonymous \texttt{anonymous\_33} (19 evaluations; E13, E18, and E19, three further candidates on E9 that were all slower, are omitted). $\ge$: truncated run. Root and tree cuts are \texttt{RootCutLevel} and \texttt{TreeCutLevel}, the two halves of \texttt{CutLevel}. Hypotheses are abridged from the rationales; the last column is our reading of the log profiles. Speedups of the selected configurations: \method 49.6$\times$ (E11), SMAC 40.3$\times$, \grimip 29.2$\times$.}
\label{tab:case}
\end{table}

\subsection{CORAL: Lowering the Node Cost}
\label{app:case:mcsched}
On the CORAL instance \texttt{mcsched}, the default run takes 47.9\,s at 12.8\,ms and 222 LP iterations per node.
Aggressive cuts (E1) raise this to 367 LP iterations per node without reducing the nodes; barrier (E3) returns the printed search path of the default run.
The agent concluded that cut generation dominates the node cost and set \texttt{CutLevel}=1 (E5): the node cost falls to 1.77\,ms, the tree grows from 3,625 to 12,702 nodes, and the solving time drops to 23.0\,s, a speedup of 2.08$\times$.
\grimip set all six of its parameters in each of its seven evaluations; none was faster than the default run, and it returned the default configuration.
\subsection{Distributions of Parameter Changes}
\label{app:case:distributions}
\begin{figure}[p]
\centering
\includegraphics[width=\textwidth]{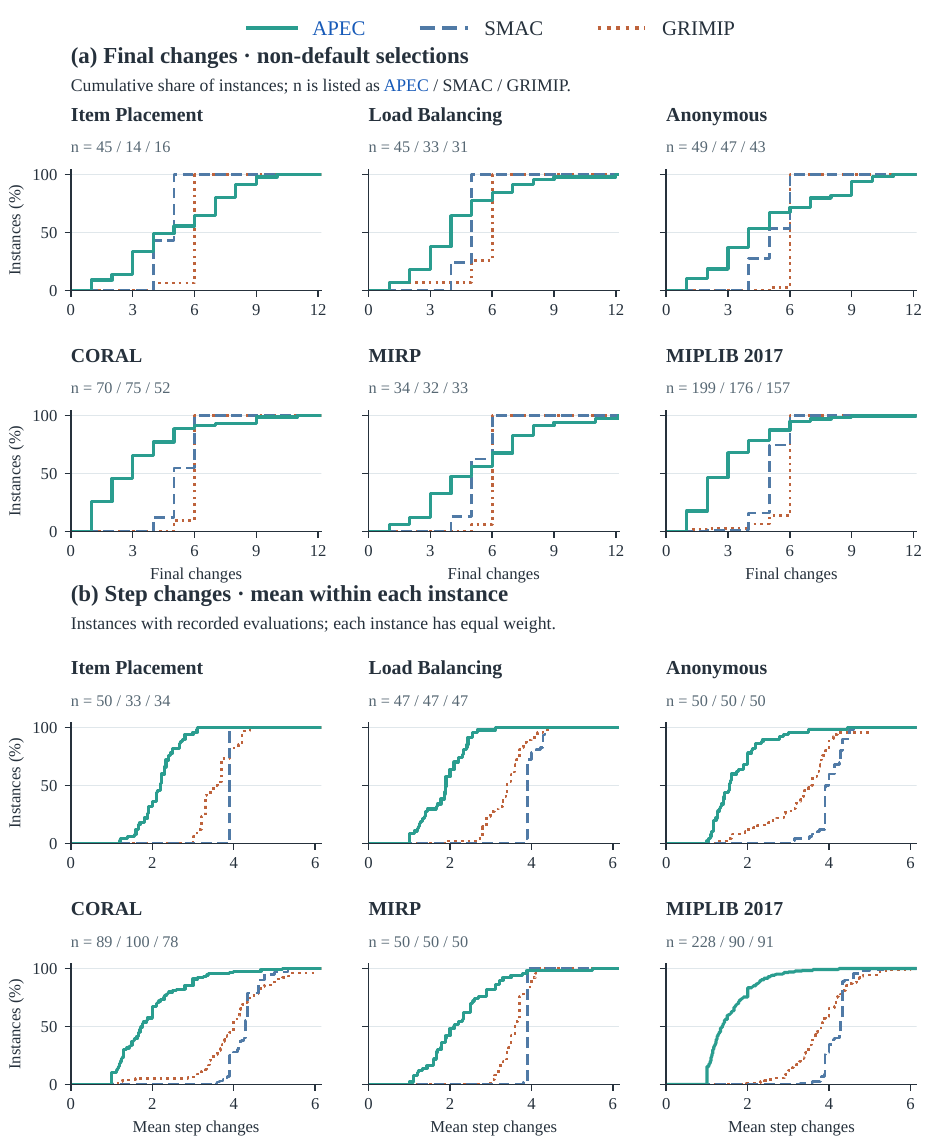}
\caption{Empirical cumulative distributions of the counts underlying Par.\ and Step in \cref{tab:main}, using the same recorded results. Curves further left indicate fewer parameter changes. (a) Final changes among non-default selections. (b) Mean step changes within each instance with recorded evaluations; instances have equal weight. The sample counts are listed in the order \method / SMAC / \grimip and may differ between methods. These counts measure readability, not explanation correctness.}
\label{fig:change-distributions}
\end{figure}

\Cref{fig:change-distributions} places the two trajectories alongside the distributions of final changes and step changes underlying \cref{tab:main}.
Each curve gives the fraction of included instances whose count is at most the value on the horizontal axis.
Final changes includes only non-default selections, as in Par.; step changes first averages within each instance and includes every instance with recorded evaluations, as in Step.
The instance sets can therefore differ between methods and between panels; each panel reports its sample counts.

The median step changes of \method is lower than that of both baselines on each of the six datasets.
In each dataset, some configurations selected by \method change more parameters than any selected by either baseline, despite its lower mean final changes on four datasets.
These distributions describe the readability of configurations and searches; they do not establish how often the explanations illustrated in the case studies are correct.
The verdict audit below addresses this separate question.

\subsection{Audit of Agent Verdicts}
\label{app:verdict-audit}
We audit the frozen results underlying \cref{tab:main}, separating recorded verdicts from their support in the evidence.
Of the 533 sessions, 358 contain instructions to record confirmed, refuted, or unexplained in the next rationale; these sessions contain 3,023 evaluations on 339 instances.
The remaining 175 sessions do not require this recording protocol and are excluded from verdict frequencies.
We link each evaluation's prediction and evaluated configuration to the verdict in the next evaluation's rationale.
We retain qualifications such as ``partially confirmed,'' distinguish competing labels, and do not infer a verdict from performance.
This channel leaves the last evaluation of each trajectory unobserved; free-form terminal explanations are outside the census.

\paragraph{Recorded verdicts.}
Of the 2,684 evaluations with a following rationale, 2,505 (93.3\%) have a linked primary label: 750 confirmed, 1,582 refuted, and 173 unexplained (\cref{tab:verdict-audit}).
The other 179 comprise 138 without an extracted label, 33 with competing labels, and eight whose verdict refers to another evaluation.
A confirmed label is the agent's assessment, and can concern only part of a prediction; its frequency is not an estimate of explanation correctness.

\begin{table}[htbp]
\centering
\small
\setlength{\tabcolsep}{5pt}
\begin{tabular}{@{}lrrrrrr@{}}
\toprule
\textbf{Dataset} & \textbf{Sessions} & \textbf{Confirmed} & \textbf{Refuted} & \textbf{Unexplained} & \textbf{Other} & \textbf{Selection} \\
\midrule
IP        & 39 & 85 & 219 & 15 & 12 & 7/36 \\
LB        & 40 & 98 & 209 & 27 & 12 & 11/38 \\
Anonymous & 42 & 144 & 263 & 21 & 17 & 8/41 \\
CORAL     & 77 & 105 & 204 & 18 & 38 & 3/48 \\
MIRP      & 42 & 107 & 219 & 19 & 26 & 2/27 \\
MIPLIB    & 118 & 211 & 468 & 73 & 74 & 25/84 \\
\midrule
Total     & 358 & 750 & 1,582 & 173 & 179 & 56/274 \\
\bottomrule
\end{tabular}
\caption{Verdict audit of the frozen results. Sessions is the number with the three-verdict recording protocol, including 19 with no evaluations. Verdict columns cover evaluations with a following rationale; Other denotes missing, competing, or unlinked labels. Selection counts non-default selections whose recorded base chain includes an observed gain labeled unexplained, divided by all non-default selections in these sessions (\cref{app:unexplained}). These are agent labels, not independently verified mechanisms.}
\label{tab:verdict-audit}
\end{table}

\paragraph{Checking the explanations.}
Experts separately review 60 verdicts from 47 instances: ten per dataset, chosen by a fixed hash ordering of the dataset, instance, and evaluation identifiers among linked primary labels, without selecting on performance or final selection.
The review compares the original prediction, actual parameter change, stated base and explicitly cited comparison runs, log profiles, and raw logs with the primary verdict and its attached explanation.
An \emph{audit judgment} is supported if the qualified verdict follows the observable prediction; contradicted if a material claim conflicts with the evidence; and insufficient if attribution exceeds the comparison.
Support does not prove an internal solver mechanism.

The audit judges 45 of the 60 explanations supported, nine contradicted, and six insufficient.
For example, on CORAL \texttt{neos-1200887}, E2 is labeled confirmed after predicting cheaper nodes, although node cost rises from 0.338 to 0.784\,ms; the faster run instead processes fewer nodes.
On MIRP \texttt{LR1\_2\_DR1\_3\_VC2\_V6a\_t60}, E7 calls a bound of $-30{,}116$ worse than $-30{,}174$ at 3,000 nodes, reversing the ordering for minimization.
Insufficient explanations attribute small time differences on identical printed search paths to LP work, or joint effects to parameters whose contributions were not isolated.
Equal dataset allocation and repeated instances make 45/60 a sample description, not a population estimate.

\paragraph{How typical are the two cases?}
Their recorded gains are strong: \texttt{anonymous\_33} has the largest speedup among the 50 Anonymous instances, while \texttt{mcsched} ranks 25th among 100 CORAL instances.
Neither uses the explicit-verdict protocol.
They illustrate a compound change and a tradeoff between node volume and node cost, but their favorable performance and the audit's counterexamples preclude treating them as evidence of uniformly reliable explanations.

\clearpage
\input{sections/miplib_trajectories}

\clearpage
\section{Solver Log Format}
\label{app:log}
A COPT~8.1 log~\citep{ge2022copt} has three parts.
The header reports the instance size before and after presolve and the ranges of the matrix, right-hand-side, bound, and cost coefficients.
The progress table follows, with one progress row printed at solver-controlled intervals; each progress row records the numbers of processed and active nodes, LP iterations per node, the number of integer-infeasible variables, the best bound, the best solution, the gap, and the elapsed time.
A leading \texttt{H} or \texttt{*} in a progress row marks a new incumbent found by a heuristic or at a node, respectively.
The footer reports the best solution, best bound, and gap, the solving time, the number of nodes, and the termination status.
\Cref{fig:example_log} shows the log of the default run on the CORAL instance \texttt{neos-1330635}.
\Cref{fig:log-to-profile} follows a longer run from its progress rows to the log profile and illustrates the provenance and comparability labels.

\begin{figure*}[t]
\centering
\footnotesize
\begin{verbatim}
Cardinal Optimizer v8.1.0-beta. Build date Jul 31 2026
[...]
Minimizing a MIP problem

The original problem has:
    2717 rows, 1736 columns and 8260 non-zero elements
    1652 binaries

Starting the MIP solver with 8 threads and 32 tasks

Presolving the problem

The presolved problem has:
    1009 rows, 1596 columns and 5516 non-zero elements
    1596 binaries

Problem info:
    Range of matrix coefficients:    [1e+00,1e+00]
    Range of rhs coefficients:       [1e+00,7e+00]
    Range of bound coefficients:     [1e+00,1e+00]
    Range of cost coefficients:      [2e+00,2e+01]
    Density of cost:                     1.8%

     Nodes    Active  LPit/n  IntInf     BestBound  BestSolution     Gap   Time
         0         1      --       0  0.000000e+00            --     Inf  0.07s
H        0         1      --       0  0.000000e+00  6.240770e+01 100.00%  0.08s
*        0         1      --       0  2.587810e+01  2.587810e+01  0.000%  0.12s
         0         1      --       0  2.587810e+01  2.587810e+01  0.000%  0.12s
         1         0     0.0       0  2.587810e+01  2.587810e+01  0.000%  0.12s
         1         0     0.0       0  2.587810e+01  2.587810e+01  0.000%  0.12s

Best solution   : 25.878100000
Best bound      : 25.878100000
Best gap        : 0.0000%
Solve time      : 0.12
Solve node      : 1
MIP status      : solved
Solution status : integer optimal (relative gap limit 0.0001)
[...]
\end{verbatim}
\caption{COPT~8.1 log of the default run on \texttt{neos-1330635} (CORAL). Lines marked \texttt{[...]} are omitted: the lines that set parameters, the hardware information, and the violation report.}
\label{fig:example_log}
\end{figure*}

\section{What an Agent Does with the Raw Log}
\label{app:study}

Before designing \method, we gave an LLM agent the raw log and watched what it did with it.
This appendix reports that trajectory study; each of its findings becomes a requirement for \method (\cref{sec:method}).

\paragraph{Pilot agent and trajectories.}
The pilot agent tuned COPT on one instance at a time.
After every evaluation it received the speedup and could open the de-identified log of the run; it had no profiles.
It ran under an earlier protocol in which each evaluation solved five permuted copies of the instance.
We collected 699 trajectories on 150 instances.
We read 283 trajectories in full, searched the remaining 416, and used LLM-written summaries of each trajectory only as leads: every case reported below was checked against the raw log or the table of evaluations.
The pilot agent opened 0.95--1.19 logs per evaluation, so it did use the log; the question is how.

\begin{table*}[t]
\centering
\small
\begin{tabularx}{\textwidth}{@{}lXX@{}}
\toprule
\textbf{Reading error} & \textbf{What the pilot agent concluded} & \textbf{What the log shows} \\
\midrule
Column confusion & 1896 is the objective of the incumbent & 1896 is in the \texttt{IntInf} column; \texttt{BestSolution} is 0 \\
First incumbent & first feasible solution at 166\,s (the first \texttt{H} row it found) & a finite incumbent already at 5.67\,s \\
Row time as cost & heuristics cost 16.86\,s & \texttt{H} rows span 1.34--18.20\,s; heuristic time is never printed \\
Large gap as weak bound & bound is weak (gap 99.82\%) & the bound reaches the optimal value at 1.00\,s; an optimal incumbent arrives at 27.95\,s \\
Printed zero gap as done & converged at 20.96\,s (gap 0.000\%) & the run ends at 90.22\,s \\
Nodes without cost & fewer nodes, so faster (488 $\rightarrow$ 9 nodes) & 89.92\,s $\rightarrow$ 122.86\,s; each node became far more expensive \\
Truncated run as finished & the tree shrank (11.6M vs.\ 31.5M nodes) & the 11.6M nodes are counted at the time limit \\
Truncated run as empty & no conclusion from a truncated run & still running at 75.12\,s, against a configuration that finished in 32.56\,s: certainly slower \\
\bottomrule
\end{tabularx}
\caption{Reading errors of the pilot agent, one representative case each, checked against the raw log. Without profiles, the agent computes every quantity itself from progress rows.}
\label{tab:reading_errors}
\end{table*}

\paragraph{Finding 1: the agent misreads the log.}
\Cref{tab:reading_errors} lists the recurring reading errors.
Some are errors of parsing, such as taking the \texttt{IntInf} column for the objective or a placeholder such as \texttt{--} for a value; others are errors of meaning, such as reading a large gap as a weak bound or fewer nodes as a faster run.
Truncated runs are misread in both directions: compared as if they had finished, or discarded although they already prove the candidate slower.
None of these errors requires deep knowledge of MIP; they arise because the agent recomputes every quantity from hundreds of progress rows, differently from one evaluation to the next.

\begin{figure}[htbp]
\centering
\includegraphics[width=\textwidth]{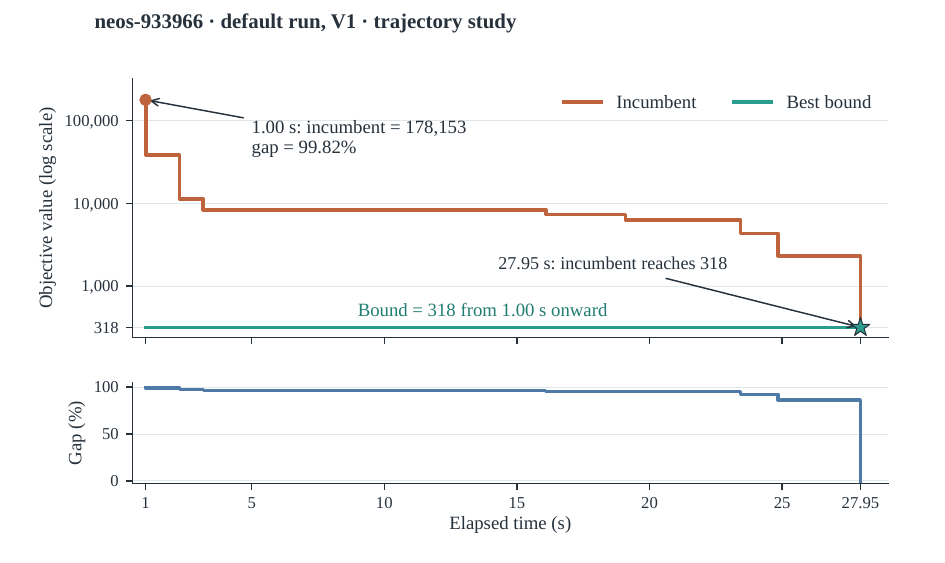}
\caption{A large gap with an already sufficient bound: the default run on V1 of \texttt{neos-933966} in the trajectory study. Curves connect printed progress rows from 1.00\,s onward using steps; the upper axis is logarithmic. The first incumbent appears earlier, at 0.28\,s. The bound is constant at 318 throughout the displayed interval; subsequent gap reduction comes from incumbent improvement. Equality with the optimal objective is established retrospectively by the completed run.}
\label{fig:gap-misreading}
\end{figure}

\Cref{fig:gap-misreading} illustrates why a large gap alone cannot identify a weak bound.
On \texttt{neos-933966}, the pilot agent proposed stronger cuts because ``the root LP bound must climb to meet the incumbent''.
In the default run on V1, however, the bound already equals the eventual optimal objective of 318 at 1.00\,s, while the incumbent is 178,153 and the gap is 99.82\%.
The incumbent reaches 318 only at 27.95\,s.
This retrospective comparison contradicts the stated bound-based rationale; it does not measure the time spent in heuristics or predict whether a cut change would help.

\paragraph{Finding 2: the agent asks the log for what it does not contain.}
\Cref{tab:log_availability} lists what an expert would want to know about a run and how much of it a COPT log provides, checked on 87,730 historical logs.
About half is unavailable: the log prints no cut counts, heuristic times, root LP iterations, or restarts.
The pilot agent nevertheless reasons about them.
It counts node-0 progress rows as cut rounds, attributes the time between two \texttt{H} rows to heuristics, and dates a restart from a drop in the number of active nodes, an event the log never prints.
An agent that is not told what the log cannot show fills the gap with plausible inventions.

\begin{table}[t]
\centering
\small
\setlength{\tabcolsep}{3pt}
\begin{tabularx}{\columnwidth}{@{}Xll@{}}
\toprule
\textbf{Quantity} & \textbf{Availability} & \textbf{Source} \\
\midrule
Presolve reduction & exact & sizes before/after \\
Presolve time & proxy & first progress row \\
Root LP time & proxy & first rows at node 0 \\
Root LP iterations & unavailable & -- \\
Number of cuts & unavailable & -- \\
Cut effect & indirect & root bound movement \\
Incumbent quality, timing & exact & \texttt{H}/\texttt{*} rows \\
Heuristic time & unavailable & -- \\
Node cost & proxy & tree time / nodes, \texttt{LPit/n} \\
Restarts & unavailable & -- \\
\bottomrule
\end{tabularx}
\caption{Diagnostic quantities and their availability in COPT logs (\texttt{Logging=1}), checked on 87,730 historical logs.}
\label{tab:log_availability}
\end{table}

\paragraph{Finding 3: a correct reading does not determine the change.}
Even a correct reading tempted the agent to turn one observation into a rule: many nodes, so add cuts; a slow presolve, so lighten it.
The trajectories contradict such rules.
Lighter presolve brought the first progress row from 20.95\,s to 6.94\,s on one instance and delayed it from 1.28\,s to 3.18\,s on another; stronger heuristics cut one run from 2,868.79\,s to about 40\,s and slowed three others; disabling cuts gave a large speedup on one instance and left another unsolved within the time limit.
Within one instance, a change that reduced the node count from 286 to 162 increased the solving time from 13.03\,s to 16.14\,s.
A bottleneck is a fact about one run; the effect of a change is a counterfactual that crosses phases and is blurred by performance variability.
The log can rank directions and rule some out, but only the next evaluation can decide.

\paragraph{Requirements.}
\textbf{R1}: quantities in the log are computed once, deterministically, and labelled by how far they can be trusted and compared, so that the agent reads facts rather than progress rows.
\textbf{R2}: what the log cannot show is stated explicitly, so that it is tested by changes rather than inferred.
\textbf{R3}: a reading leads to a hypothesis that names the evidence that would refute it, and the next evaluation is designed as that test.
R1 and R2 are what a profiler provides in GPU kernel optimisation; R3 is what the engineer does with its output.

\section{Adapting DeepSeek Harness for Solver Tuning}
\label{app:dsh}
\method uses a fork of DeepSeek Harness (dsh)\footnote{\url{https://github.com/deepseek-ai/deepseek-harness}} to execute the agent loop.
We build on dsh~0.2.0-rc.2 (upstream commit \texttt{639ed01539}) and distribute the adaptation as a patch, producing commit \texttt{a01a04b7f3}.
The adaptation connects dsh to the Python implementation of the three layers in \cref{sec:method}; the solver and the decisions governing evaluations remain on the Python side.

\paragraph{A dedicated tuning profile.}
We add a standalone profile that retains dsh's headless runner, agent loop, DeepSeek adapter, retries, token accounting, context compaction, and JSONL session records.
The profile omits the general coding tools, telemetry, session-title generation, and automatic uploads of session logs and plugin inventories.
The Python driver supplies the system prompt and tool schemas through a per-session manifest.
The skill registry and the \texttt{skill} and \texttt{bash} tools are mounted only when enabled for that session; skill discovery uses explicitly supplied directories rather than the host's default directories.

\paragraph{Tools and control through a bridge.}
A new dsh plugin registers the manifest's tuning tools and forwards their calls to the Python driver over an authenticated loopback HTTP bridge.
Only tools allowed by the manifest can execute, and a driver response ending the session blocks subsequent tool calls.
Before each agent step, including any context compaction, the plugin asks the driver whether the session may continue; the driver enforces the budget and step limits and can send a control message.
When the agent replies without a tool call, the driver can request that it continue tuning; the system ends the session when the remaining budget cannot fund another evaluation or model step.
The plugin also measures LLM calls, including compaction, and reports their duration and token usage to the driver.
For skill and shell calls, it checks permission with the driver before execution and reports elapsed time afterward.
These hooks let the system layer charge analysis to the same budget as solver evaluations and enforce the protocol independently of the agent's text.

\paragraph{Compaction with an exact evaluation record.}
We extend dsh's basic compaction engine to accept a tuning-specific instruction supplied by the driver.
The compacted context contains the LLM's working notes followed by the driver's exact session record, so configurations and evaluation results need not be reconstructed from an LLM-generated summary.
This requires exporting the upstream summarization helper and allowing a custom instruction; token accounting and retention of recent context reuse dsh's existing machinery.

\paragraph{De-identification and shell isolation.}
We add a hook to the DeepSeek adapter immediately before an HTTP request is sent.
The driver applies de-identification to string values throughout the serialized request, including compaction requests, and archives the resulting request body; if this step fails, the request is not sent.
We also extend dsh's local shell executor to accept a sandbox launcher and map working directories into the sandbox.
The Python driver constructs the bubblewrap environment: de-identified profiles and logs and the skill resources are read-only, scratch space is writable, and the instance file, solver licence, host network, and API credentials are unavailable to shell commands.
Thus, the agent can inspect evidence and run analysis scripts while solver evaluations remain under the driver's control.

\paragraph{Reproducible session setup.}
Each session receives its own dsh home directory, manifest, and configuration patch.
The process starts in a neutral temporary directory to keep instance-identifying paths out of session metadata; API credentials are passed through the process environment.
We retain the JSONL session record, the manifest, the configuration patch, and the de-identified LLM request bodies for inspection.
These adaptations implement the execution requirements of \method; we do not attribute tuning gains to the dsh changes themselves.

\section{Dataset Coverage and Sample Counts}
\label{app:dataset_counts}
\Cref{tab:main} compares all methods on the same 533 instances.
The Gap subset contains only instances whose default run reaches the time limit.

\paragraph{MIPLIB~2017 benchmark coverage.}
The dataset is the full MIPLIB~2017 benchmark set of 240 instances~\citep{MIPLIB2017}, including seven known infeasible instances.\footnote{Official benchmark list: \url{https://miplib.zib.de/tag_benchmark.html}.}
COPT proves infeasibility in the default runs for \texttt{bnatt500}, \texttt{fhnw-binpack4-4}, \texttt{neos859080}, and \texttt{neos-3988577-wolgan}, in agreement with the official benchmark.
Proving infeasibility is a successful solver outcome.
However, the tuning implementation used for these experiments classified these completed runs as failures and did not retain them in the default cache.
The main results and \cref{fig:teaser} therefore aggregate the remaining 236 instances, shared by all methods.
The other three known infeasible instances, \texttt{neos-3402454-bohle}, \texttt{neos-2075418-temuka}, and \texttt{cryptanalysiskb128n5obj14}, remain in this set: their default runs reach the time limit without proving infeasibility.

\begin{table}[ht]
\centering
\small
\begin{tabular}{lrr}
\toprule
\textbf{Dataset} & \textbf{Evaluated instances} & \textbf{Gap subset} \\
\midrule
Item Placement & 50 & 50 \\
Load Balancing & 47 & 33 \\
Anonymous & 50 & 25 \\
CORAL & 100 & 20 \\
MIRP & 50 & 49 \\
MIPLIB~2017 & 236 & 34 \\
\bottomrule
\end{tabular}
\caption{Instance counts for the main results.}
\label{tab:dataset_counts}
\end{table}

\paragraph{Random coverage.}
Random is reported on the same 533 instances.
Verified recommendations are available for 506 instances: 223 select the default configuration, 280 have complete measurements of a non-default configuration, and 3 lack a complete measurement.
The remaining 27 instances have no verified recommendation.
Random uses the default configuration whenever a verified recommendation or complete measurement is unavailable.

Par.\ is averaged over each tuner's instances with a non-default selection (\cref{tab:par_counts}); the included instances can differ between tuners.

\begin{table}[ht]
\centering
\small
\setlength{\tabcolsep}{4pt}
\begin{tabular}{lrrrr}
\toprule
\textbf{Dataset} & \textbf{\method} & \textbf{Random} & \textbf{SMAC} & \textbf{\grimip} \\
\midrule
Item Placement & 45 & 22 & 14 & 16 \\
Load Balancing & 45 & 11 & 33 & 31 \\
Anonymous & 49 & 31 & 47 & 43 \\
CORAL & 70 & 57 & 75 & 52 \\
MIRP & 34 & 10 & 32 & 33 \\
MIPLIB~2017 & 199 & 133 & 176 & 157 \\
\bottomrule
\end{tabular}
\caption{Instances with a non-default selection included in Par.}
\label{tab:par_counts}
\end{table}

Step is averaged over instances with recorded evaluation configurations, including instances on which the selected configuration is the default configuration.
For \method, Step includes 89 CORAL instances and 228 MIPLIB~2017 instances; all instances are included on the other datasets.
For Random, complete evaluation sequences are available for all instances in Item Placement, Load Balancing, Anonymous, CORAL, and MIRP, and for 214 MIPLIB~2017 instances.
Step can be computed from a complete evaluation sequence even when its final recommendation is unavailable.
For SMAC and \grimip, respectively, Step on Item Placement is averaged over 33 and 34 instances with recorded evaluation configurations; on MIPLIB~2017, it is averaged over 90 and 91 instances.

\paragraph{Ablation coverage.}
The default-time criterion $t(c_0)\le600$\,s identifies 191 MIPLIB~2017 benchmark instances, including the four instances whose default runs prove infeasibility.
The ablations in \cref{tab:ablation} use the remaining 187 instances with completed \method runs and reusable default runs.
Par.\ includes 158 instances for \method, 141 for the raw-log variant, and 151 for w/o interpretation layer.
Using a factor of 1.02 to compare solving times with \method in either direction, the raw-log variant selects a faster configuration on 57 instances and a slower one on 86.
The \emph{w/o interpretation layer} variant selects a faster configuration on 68 instances and a slower one on 79.

\section{Metric Calculation}
\label{app:metrics}
For solving times $t_1,\ldots,t_n$, the shifted geometric mean is
\[
\mathrm{SGM10}(t)=\exp\!\left(\frac{1}{n}\sum_{i=1}^{n}\log(t_i+10)\right)-10.
\]
Imp.\ divides this quantity for Default by the same quantity for the selected configurations on the same instances; Default therefore scores $1.00\times$.
For Eff., improvement means a shorter solving time when Default solves the instance, or solving the instance or obtaining a smaller final gap when Default times out.
The Gap subset is fixed by Default and shared by all tuners.
Gap uses the analogous shifted geometric mean with a shift of 1 percentage point, after capping each final relative gap at 100\% and assigning 100\% to runs without an incumbent.
Gap supplements the solving-time metrics and does not change the tuning objective or final selection.
For each tuner, Par.\ is the arithmetic mean of final changes over instances whose selected configuration differs from the default configuration.
Selections equal to the default configuration are excluded from both the sum and the instance count; if no non-default configuration is selected, Par.\ is reported as --.
Step refers to the whole search, excluding the default run and candidates rejected before evaluation (\cref{sec:interp:measures}).

\paragraph{Best Observed.}
For each instance, Best Observed takes the fastest solved outcome, or, if none is solved, the outcome with the smallest final gap among runs reaching the full time limit; Default wins ties.
It pools results from multiple tuning budgets.

\subsection{Budget Curves}
\label{app:budget-curves}
\Cref{fig:budget-curves} uses the 209 MIPLIB~2017 instances with complete, verified Random results and available trajectories for all four tuners.
Of the other 27 instances in the main-result set, the available Random records contain 19 interrupted searches, five requiring verification, and three without a search record.
The same 209 instances are used for every curve, Default, and Best Observed.

\paragraph{Budget and presentation.}
The normalized consumed budget $u\in[0,1]$ divides the cumulative charged time at the end of an evaluation by that trajectory's budget $10t(c_0)$.
The recorded charges include solver, LLM, and tool time.
For \method, each budget prefix applies final selection to eligible evaluations completed by that point, retaining Default when no candidate qualifies.
Failed evaluations, truncated runs, and evaluations finishing beyond the recorded budget cannot improve its prefix.
The solid line shows \method; dashed lines show baselines.
Circles mark final selections.

\paragraph{Aggregation and uncertainty.}
Performance relative to Best Observed is
\[
  R_m(u)=\frac{\mathrm{SGM10}(\{t_i^{\mathrm{Best\ Observed}}\})}{G_m(u)}.
\]
Here $G_m(u)$ denotes the SGM10 value for method $m$ at budget $u$.
SGM10 is computed on times in seconds before taking the ratio; Best Observed is therefore 100\%.
We use 2,000 paired bootstrap samples of instances, sharing instance weights across tuners and Best Observed.
Shading is the plotted value plus or minus one bootstrap standard deviation.
It describes variation under the bootstrap procedure rather than variation across repeated searches.

\section{Profile Fields}
\label{app:fields}
\Cref{tab:log_fields_a,tab:log_fields_b,tab:log_fields_c} list the 126 fields of the log profile and \cref{tab:structure_fields} the 52 fields of the structure profile, group by group; fields that share a meaning and labels share a row.
Times are in seconds and gaps in percent, as COPT prints them; fields ending in \texttt{\_frac} are fractions of the observed run unless stated otherwise, and fields ending in \texttt{\_steps} count objective steps, computed only when the step is known and the printed precision resolves it.
A missing value is reported as missing, never as zero.

The structure profile describes the instance file, which does not change with the configuration, so it carries no comparability label.
Its provenance label is \emph{exact}, counted from the file; \emph{sufficient}, asserted only when proven, so that a negative value means ``not proven''; or \emph{proxy}, a convention with a fixed threshold.

\begin{figure}[htbp]
\centering
\includegraphics[width=\textwidth]{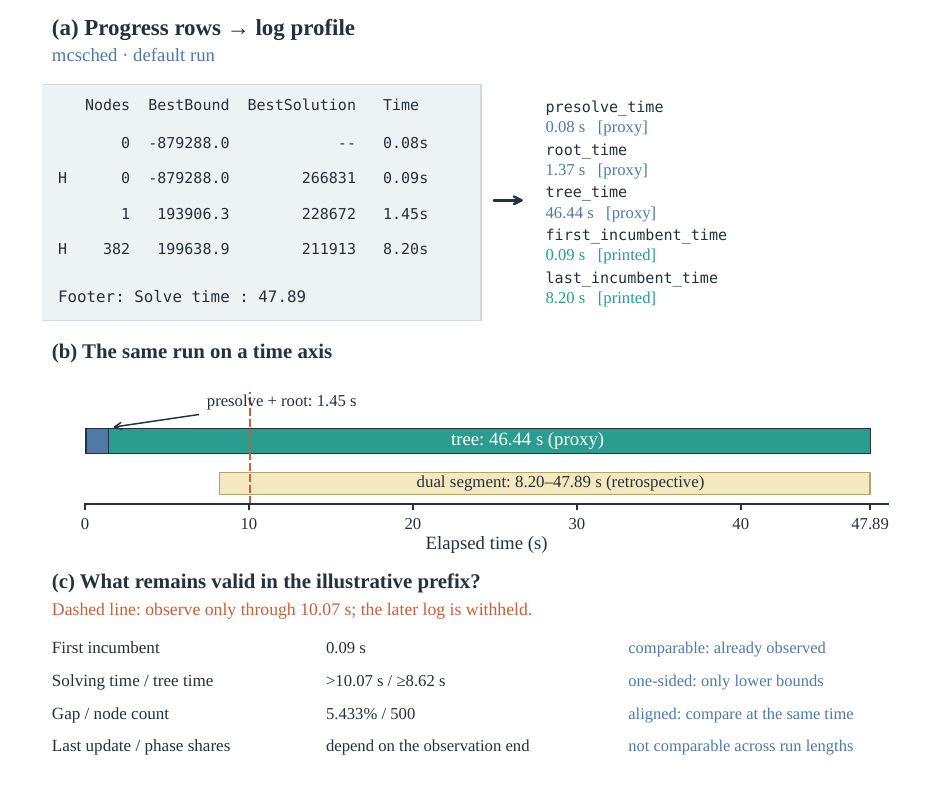}
\caption{From a solver log to a log profile on \texttt{mcsched} (CORAL). (a) Selected columns and rows, with objective values reformatted for readability, and the resulting fields with provenance labels. (b) Phase times are progress-row proxies; the dual segment is identified retrospectively. (c) An illustrative prefix of the same log ends at 10.07\,s, without a footer. This prefix is not a separate solver run. Values already observed, one-sided bounds, comparisons requiring alignment, and quantities that depend on the observation end must be treated differently.}
\label{fig:log-to-profile}
\end{figure}

\Cref{fig:log-to-profile} gives a concrete reading of these labels for the log profile.
In the default run on \texttt{mcsched}, the first progress row is at 0.08\,s and the first tree row at 1.45\,s; the footer reports 47.89\,s.
These timestamps give presolve, root, and tree proxies of 0.08, 1.37, and 46.44\,s.
Withholding the log after the progress row at 10.07\,s leaves the first incumbent observable, but only lower bounds on completed solving time and tree time.
The last incumbent update in a prefix need not be the last update of the complete run.

\begin{table*}[p]
\centering
\footnotesize
\setlength{\tabcolsep}{3pt}
\renewcommand{\arraystretch}{1.0}
\begin{tabularx}{\textwidth}{@{}>{\raggedright\arraybackslash\hsize=1.2\hsize}X>{\raggedright\arraybackslash\hsize=0.8\hsize}Xll@{}}
\toprule
\textbf{Fields} & \textbf{Meaning} & \textbf{Prov.} & \textbf{Comp.} \\
\midrule
\multicolumn{4}{@{}l}{\textbf{run} (18): can this log be trusted and compared?} \\
\texttt{censored} & the log does not end as solved, infeasible, or unbounded & printed & -- \\
\texttt{terminal\_\allowbreak{}complete} & the footer prints MIP status, solving time, and node count & printed & -- \\
\texttt{time\_\allowbreak{}basis}, \texttt{observed\_\allowbreak{}until} & end of the observed run: footer time, or the last progress row if the footer is missing & printed & -- \\
\texttt{progress\_\allowbreak{}rows}, \texttt{progress\_\allowbreak{}monotone} & number of parsed progress rows; whether time and node counts never decrease & printed & -- \\
\texttt{phase\_\allowbreak{}timing\_\allowbreak{}quality} & whether the \emph{phases} group can be used & proxy & -- \\
\texttt{primal\_\allowbreak{}values\_\allowbreak{}available}, \texttt{dual\_\allowbreak{}values\_\allowbreak{}available} & whether finite incumbent / bound values are printed & printed & -- \\
\texttt{incumbent\_\allowbreak{}event\_\allowbreak{}detection} & incumbent updates detected from printed value changes, or unavailable & printed & -- \\
\texttt{time\_\allowbreak{}limit}, \texttt{threads}, \texttt{tasks}, \texttt{rel\_\allowbreak{}gap}, \texttt{solver\_\allowbreak{}version} & run conditions printed in the log & printed & -- \\
\texttt{warnings} & non-identifying solver warnings & printed & -- \\
\texttt{numerical\_\allowbreak{}issue\_\allowbreak{}warning\_\allowbreak{}lines}, \texttt{numerical\_\allowbreak{}issues\_\allowbreak{}reported} & visible numerical-issue rows; LP numerical issues counted in the footer & printed & -- \\
\midrule
\multicolumn{4}{@{}l}{\textbf{outcome} (8): how did the run end?} \\
\texttt{status} & status at the end of the log (solved, time limit, interrupted, \dots) & printed & -- \\
\texttt{solve\_\allowbreak{}time}, \texttt{nodes} & solving time and node count from the footer & printed & one-sided \\
\texttt{final\_\allowbreak{}gap}, \texttt{final\_\allowbreak{}gap\_\allowbreak{}steps} & final gap, in percent and in objective steps & printed & \xmark \\
\texttt{has\_\allowbreak{}solution}, \texttt{solved\_\allowbreak{}in\_\allowbreak{}presolve} & an incumbent exists at the end; presolve solved the instance & printed & \xmark \\
\texttt{tree\_\allowbreak{}end\_\allowbreak{}mode} & unfinished, presolve, root, exhausted, or closed with open nodes & proxy & \xmark \\
\midrule
\multicolumn{4}{@{}l}{\textbf{problem} (14): what is the instance? (same for every configuration)} \\
\texttt{orig\_\allowbreak{}rows}, \texttt{orig\_\allowbreak{}cols}, \texttt{orig\_\allowbreak{}nonzeros}, \texttt{orig\_\allowbreak{}binaries}, \texttt{orig\_\allowbreak{}integers} & size of the original instance & printed & -- \\
\texttt{matrix\_\allowbreak{}range\_\allowbreak{}log10}, \texttt{rhs\_\allowbreak{}range\_\allowbreak{}log10}, \texttt{bound\_\allowbreak{}range\_\allowbreak{}log10}, \texttt{cost\_\allowbreak{}range\_\allowbreak{}log10} & orders of magnitude spanned by each coefficient range & printed & -- \\
\texttt{cost\_\allowbreak{}density\_\allowbreak{}pct}, \texttt{zero\_\allowbreak{}objective}, \texttt{objective\_\allowbreak{}sense} & objective density; zero objective; direction & printed & -- \\
\texttt{objective\_\allowbreak{}step}, \texttt{objective\_\allowbreak{}step\_\allowbreak{}source} & objective step, proven by the structure profile (sufficient) or assumed from integral incumbents & sufficient/assumed & -- \\
\midrule
\multicolumn{4}{@{}l}{\textbf{presolve} (8): how much did presolve remove?} \\
\texttt{presolved\_\allowbreak{}rows}, \texttt{presolved\_\allowbreak{}cols}, \texttt{presolved\_\allowbreak{}nonzeros}, \texttt{presolved\_\allowbreak{}binaries}, \texttt{presolved\_\allowbreak{}integers} & size after presolve & printed & \cmark \\
\texttt{presolve\_\allowbreak{}removed\_\allowbreak{}rows\_\allowbreak{}frac}, \texttt{presolve\_\allowbreak{}removed\_\allowbreak{}cols\_\allowbreak{}frac}, \texttt{presolve\_\allowbreak{}removed\_\allowbreak{}nonzeros\_\allowbreak{}frac} & fraction removed (negative if presolve adds rows or columns) & printed & \cmark \\
\bottomrule
\end{tabularx}
\caption{Log-profile fields, part 1 of 3 (\emph{run} to \emph{presolve}). Provenance: printed, proxy, retro.\ (retrospective), assumed; the objective step proven by the structure profile keeps its label there, sufficient. Comparability: \cmark{} comparable, one-sided, aligned, \xmark{} not comparable; -- marks run conditions and instance-level fields, which are not compared.}
\label{tab:log_fields_a}
\end{table*}

\begin{table*}[p]
\centering
\footnotesize
\setlength{\tabcolsep}{3pt}
\renewcommand{\arraystretch}{1.0}
\begin{tabularx}{\textwidth}{@{}>{\raggedright\arraybackslash\hsize=1.2\hsize}X>{\raggedright\arraybackslash\hsize=0.8\hsize}Xll@{}}
\toprule
\textbf{Fields} & \textbf{Meaning} & \textbf{Prov.} & \textbf{Comp.} \\
\midrule
\multicolumn{4}{@{}l}{\textbf{phases} (14): where did the time go?} \\
\texttt{presolve\_\allowbreak{}time} & start to the first node-0 progress row & proxy & \cmark \\
\texttt{root\_\allowbreak{}lp\_\allowbreak{}time} & first row to the second non-\texttt{H} node-0 row & proxy & \cmark \\
\texttt{root\_\allowbreak{}rounds\_\allowbreak{}time}, \texttt{root\_\allowbreak{}rounds\_\allowbreak{}lp\_\allowbreak{}time}, \texttt{root\_\allowbreak{}rounds\_\allowbreak{}heur\_\allowbreak{}time} & remaining node-0 rows, split by whether a row is an \texttt{H} row; not cut time & proxy & \cmark \\
\texttt{root\_\allowbreak{}silent\_\allowbreak{}time}, \texttt{root\_\allowbreak{}time} & last node-0 row to the first tree row; whole root & proxy & \cmark \\
\texttt{tree\_\allowbreak{}time} & first tree row to the end & proxy & one-sided \\
\texttt{presolve\_\allowbreak{}frac}, \texttt{root\_\allowbreak{}frac}, \texttt{tree\_\allowbreak{}frac} & the three phases as fractions of the observed run & retro. & \xmark \\
\texttt{root\_\allowbreak{}lp\_\allowbreak{}solved}, \texttt{root\_\allowbreak{}transition\_\allowbreak{}observed}, \texttt{reached\_\allowbreak{}tree} & whether the root LP boundary, the node-1 row, and a tree row were seen & proxy & -- \\
\midrule
\multicolumn{4}{@{}l}{\textbf{root} (8): how far did the root bound move?} \\
\texttt{root\_\allowbreak{}rounds} & non-\texttt{H} node-0 rows after the root LP; not cut rounds & proxy & \cmark \\
\texttt{root\_\allowbreak{}heuristic\_\allowbreak{}rows} & \texttt{H} node-0 rows & printed & \cmark \\
\texttt{root\_\allowbreak{}lp\_\allowbreak{}intinf}, \texttt{root\_\allowbreak{}final\_\allowbreak{}intinf}, \texttt{root\_\allowbreak{}end\_\allowbreak{}gap} & \texttt{IntInf} after the root LP and at the end of the root; gap at the end of the root & printed & \cmark \\
\texttt{root\_\allowbreak{}gap\_\allowbreak{}closed} & root bound movement over the distance to the final incumbent & retro. & \xmark \\
\texttt{root\_\allowbreak{}share\_\allowbreak{}of\_\allowbreak{}bound\_\allowbreak{}progress} & share of total bound movement made at the root & retro. & \xmark \\
\texttt{root\_\allowbreak{}bound\_\allowbreak{}deficit\_\allowbreak{}steps} & steps from the last root bound to the final incumbent & retro. & \xmark \\
\midrule
\multicolumn{4}{@{}l}{\textbf{primal} (29): when were incumbents found, and how long was the dual segment?} \\
\texttt{first\_\allowbreak{}incumbent\_\allowbreak{}time}, \texttt{first\_\allowbreak{}incumbent\_\allowbreak{}node}, \texttt{first\_\allowbreak{}incumbent\_\allowbreak{}source}, \texttt{first\_\allowbreak{}incumbent\_\allowbreak{}phase}, \texttt{first\_\allowbreak{}incumbent\_\allowbreak{}gap} & time, node, source (\texttt{H}, \texttt{*}, other), phase, and gap of the first incumbent & printed & \cmark \\
\texttt{first\_\allowbreak{}incumbent\_\allowbreak{}frac} & time to first incumbent as a fraction of the observed run & retro. & \xmark \\
\texttt{first\_\allowbreak{}heuristic\_\allowbreak{}solution\_\allowbreak{}time}, \texttt{first\_\allowbreak{}node\_\allowbreak{}solution\_\allowbreak{}time} & first \texttt{H} and \texttt{*} rows, distinct from the first incumbent & printed & \cmark \\
\texttt{last\_\allowbreak{}incumbent\_\allowbreak{}time}, \texttt{last\_\allowbreak{}incumbent\_\allowbreak{}node}, \texttt{last\_\allowbreak{}incumbent\_\allowbreak{}source}, \texttt{last\_\allowbreak{}incumbent\_\allowbreak{}phase}, \texttt{last\_\allowbreak{}incumbent\_\allowbreak{}gap} & the last printed incumbent update & printed & \xmark \\
\texttt{last\_\allowbreak{}incumbent\_\allowbreak{}frac}, \texttt{last\_\allowbreak{}incumbent\_\allowbreak{}gap\_\allowbreak{}steps} & its time as a fraction; steps still to close in the dual segment & retro. & \xmark \\
\texttt{nodes\_\allowbreak{}after\_\allowbreak{}last\_\allowbreak{}incumbent\_\allowbreak{}frac}, \texttt{primal\_\allowbreak{}stall\_\allowbreak{}frac} & share of nodes and of time in the dual segment & retro. & \xmark \\
\texttt{gap\_\allowbreak{}closure\_\allowbreak{}primal\_\allowbreak{}frac} & share of the gap closed by the incumbent & retro. & \xmark \\
\texttt{incumbent\_\allowbreak{}updates}, \texttt{incumbent\_\allowbreak{}updates\_\allowbreak{}heuristic}, \texttt{incumbent\_\allowbreak{}updates\_\allowbreak{}node}, \texttt{incumbent\_\allowbreak{}updates\_\allowbreak{}root}, \texttt{incumbent\_\allowbreak{}updates\_\allowbreak{}tree}, \texttt{incumbent\_\allowbreak{}event\_\allowbreak{}time\_\allowbreak{}groups} & incumbent updates by source and phase; distinct update times & printed & \xmark \\
\texttt{reported\_\allowbreak{}solution\_\allowbreak{}events}, \texttt{reported\_\allowbreak{}solution\_\allowbreak{}events\_\allowbreak{}heuristic}, \texttt{reported\_\allowbreak{}solution\_\allowbreak{}events\_\allowbreak{}node}, \texttt{reported\_\allowbreak{}solution\_\allowbreak{}time\_\allowbreak{}groups}, \texttt{last\_\allowbreak{}reported\_\allowbreak{}solution\_\allowbreak{}time} & \texttt{H}/\texttt{*} rows, which may repeat an incumbent; distinct from incumbent updates & printed & \xmark \\
\bottomrule
\end{tabularx}
\caption{Log-profile fields, part 2 of 3 (\emph{phases} to \emph{primal}). Labels as in \cref{tab:log_fields_a}.}
\label{tab:log_fields_b}
\end{table*}

\begin{table*}[p]
\centering
\footnotesize
\setlength{\tabcolsep}{3pt}
\renewcommand{\arraystretch}{1.0}
\begin{tabularx}{\textwidth}{@{}>{\raggedright\arraybackslash\hsize=1.2\hsize}X>{\raggedright\arraybackslash\hsize=0.8\hsize}Xll@{}}
\toprule
\textbf{Fields} & \textbf{Meaning} & \textbf{Prov.} & \textbf{Comp.} \\
\midrule
\multicolumn{4}{@{}l}{\textbf{bound} (8): how did the dual bound move?} \\
\texttt{bound\_\allowbreak{}stall\_\allowbreak{}frac}, \texttt{last\_\allowbreak{}bound\_\allowbreak{}change\_\allowbreak{}frac} & time since, and time of, the last bound change, as fractions & retro. & \xmark \\
\texttt{gap\_\allowbreak{}closure\_\allowbreak{}dual\_\allowbreak{}frac} & share of the gap closed by the bound & retro. & \xmark \\
\texttt{tree\_\allowbreak{}bound\_\allowbreak{}delta\_\allowbreak{}pct}, \texttt{tree\_\allowbreak{}bound\_\allowbreak{}delta\_\allowbreak{}steps} & bound movement in the tree & retro. & \xmark \\
\texttt{final\_\allowbreak{}bound\_\allowbreak{}jump\_\allowbreak{}pct}, \texttt{final\_\allowbreak{}bound\_\allowbreak{}jump\_\allowbreak{}steps}, \texttt{final\_\allowbreak{}bound\_\allowbreak{}jump\_\allowbreak{}share} & bound movement between the last two rows and its share of the tree movement & retro. & \xmark \\
\midrule
\multicolumn{4}{@{}l}{\textbf{tree} (11): how large and costly was the tree?} \\
\texttt{tree\_\allowbreak{}nodes\_\allowbreak{}per\_\allowbreak{}sec}, \texttt{tree\_\allowbreak{}ms\_\allowbreak{}per\_\allowbreak{}node} & node throughput over the tree time & proxy & \xmark \\
\texttt{lpit\_\allowbreak{}per\_\allowbreak{}node} & \texttt{LPit/n} of the last tree row & printed & \xmark \\
\texttt{max\_\allowbreak{}active}, \texttt{final\_\allowbreak{}active}, \texttt{max\_\allowbreak{}active\_\allowbreak{}nodes} & peak and final open nodes; node count at the peak & printed & \xmark \\
\texttt{max\_\allowbreak{}active\_\allowbreak{}time\_\allowbreak{}frac}, \texttt{active\_\allowbreak{}decline\_\allowbreak{}frac} & time of the peak as a fraction; decline from the peak & retro. & \xmark \\
\texttt{tree\_\allowbreak{}window\_\allowbreak{}count}, \texttt{tree\_\allowbreak{}windows} & up to six windows between progress rows with node throughput & proxy & aligned \\
\texttt{tree\_\allowbreak{}segments} & up to six time segments with throughput, open nodes, gap, bound movement, and incumbent updates & proxy & aligned \\
\midrule
\multicolumn{4}{@{}l}{\textbf{gap\_progress} (6): when were given gaps reached?} \\
\texttt{gap\_\allowbreak{}threshold\_\allowbreak{}times}, \texttt{gap\_\allowbreak{}threshold\_\allowbreak{}nodes} & time and node count at which the gap first reaches 10, 1, 0.1, 0.01\% & printed & \cmark \\
\texttt{first\_\allowbreak{}displayed\_\allowbreak{}zero\_\allowbreak{}gap\_\allowbreak{}time} & first progress row that prints a zero gap & printed & \cmark \\
\texttt{time\_\allowbreak{}after\_\allowbreak{}displayed\_\allowbreak{}zero\_\allowbreak{}gap}, \texttt{gap\_\allowbreak{}stall\_\allowbreak{}frac}, \texttt{last\_\allowbreak{}gap\_\allowbreak{}change\_\allowbreak{}frac} & time after it; time since, and time of, the last gap change & retro. & \xmark \\
\midrule
\multicolumn{4}{@{}l}{\textbf{signature} (2): did the printed search path change?} \\
\texttt{root\_\allowbreak{}signature} & hash of the printed columns of node-0 rows, excluding time and objective values & printed & \cmark \\
\texttt{search\_\allowbreak{}signature} & the same over all progress rows, the final node count, and the status & printed & \xmark \\
\bottomrule
\end{tabularx}
\caption{Log-profile fields, part 3 of 3 (\emph{bound} to \emph{signature}). Labels as in \cref{tab:log_fields_a}.}
\label{tab:log_fields_c}
\end{table*}

\begin{table*}[p]
\centering
\footnotesize
\setlength{\tabcolsep}{3pt}
\renewcommand{\arraystretch}{1.0}
\begin{tabularx}{\textwidth}{@{}>{\raggedright\arraybackslash\hsize=1.2\hsize}X>{\raggedright\arraybackslash\hsize=0.8\hsize}Xl@{}}
\toprule
\textbf{Fields} & \textbf{Meaning} & \textbf{Provenance} \\
\midrule
\multicolumn{3}{@{}l}{\textbf{size} (3): how large is the instance?} \\
\texttt{rows}, \texttt{cols}, \texttt{nonzeros} & size of the file & exact \\
\midrule
\multicolumn{3}{@{}l}{\textbf{variables} (12): what types and domains do the variables have?} \\
\texttt{binary\_\allowbreak{}cols}, \texttt{general\_\allowbreak{}integer\_\allowbreak{}cols}, \texttt{continuous\_\allowbreak{}cols} & variable types & exact \\
\texttt{implied\_\allowbreak{}integer\_\allowbreak{}cols} & continuous variables proven integral through equality rows & sufficient \\
\texttt{fixed\_\allowbreak{}cols}, \texttt{empty\_\allowbreak{}cols} & fixed variables; variables in no row & exact \\
\texttt{general\_\allowbreak{}integer\_\allowbreak{}domain\_\allowbreak{}p50}, \texttt{general\_\allowbreak{}integer\_\allowbreak{}domain\_\allowbreak{}max}, \texttt{general\_\allowbreak{}integer\_\allowbreak{}small\_\allowbreak{}domain\_\allowbreak{}frac} & domain widths of bounded general integers; the fraction of width at most 10 is over all general integers & exact \\
\texttt{unbounded\_\allowbreak{}integer\_\allowbreak{}cols}, \texttt{free\_\allowbreak{}continuous\_\allowbreak{}cols}, \texttt{unbounded\_\allowbreak{}continuous\_\allowbreak{}cols} & unbounded and free variables & exact \\
\midrule
\multicolumn{3}{@{}l}{\textbf{objective} (5): where is the objective, and does it have a step?} \\
\texttt{objective\_\allowbreak{}sense}, \texttt{objective\_\allowbreak{}nonzeros}, \texttt{objective\_\allowbreak{}cost\_\allowbreak{}on} & direction; nonzero objective coefficients by variable type & exact \\
\texttt{objective\_\allowbreak{}integral}, \texttt{objective\_\allowbreak{}step} & objective provably integral; its step, passed to the log profile & sufficient \\
\midrule
\multicolumn{3}{@{}l}{\textbf{constraints} (9): what types of rows are there?} \\
\texttt{row\_\allowbreak{}classes}, \texttt{row\_\allowbreak{}class\_\allowbreak{}frac} & 17-class row classification adapted from MIPLIB 2017 (\cref{app:structure}) & exact \\
\texttt{equality\_\allowbreak{}rows\_\allowbreak{}frac}, \texttt{ranged\_\allowbreak{}rows}, \texttt{integral\_\allowbreak{}coefficient\_\allowbreak{}rows\_\allowbreak{}frac} & equality, ranged, and integral-coefficient rows & exact \\
\texttt{indicator\_\allowbreak{}constraints}, \texttt{sos\_\allowbreak{}constraints}, \texttt{quadratic\_\allowbreak{}nonzeros}, \texttt{quadratic\_\allowbreak{}constraints} & special constraints & exact \\
\midrule
\multicolumn{3}{@{}l}{\textbf{numerics} (5): are coefficients badly scaled; are there big-M rows?} \\
\texttt{bigm\_\allowbreak{}rows}, \texttt{bigm\_\allowbreak{}ratio\_\allowbreak{}log10\_\allowbreak{}max} & mixed rows whose largest integer/continuous coefficient ratio is at least $10^3$; the largest ratio over all mixed rows & proxy \\
\texttt{row\_\allowbreak{}dynamism\_\allowbreak{}log10\_\allowbreak{}p50}, \texttt{row\_\allowbreak{}dynamism\_\allowbreak{}log10\_\allowbreak{}p90}, \texttt{row\_\allowbreak{}dynamism\_\allowbreak{}log10\_\allowbreak{}max} & spread of coefficient magnitudes within a row & exact \\
\midrule
\multicolumn{3}{@{}l}{\textbf{sparsity} (5): how dense are rows and columns?} \\
\texttt{row\_\allowbreak{}nnz\_\allowbreak{}p50}, \texttt{row\_\allowbreak{}nnz\_\allowbreak{}max}, \texttt{col\_\allowbreak{}nnz\_\allowbreak{}p50}, \texttt{col\_\allowbreak{}nnz\_\allowbreak{}max} & nonzeros per row and per column & exact \\
\texttt{dense\_\allowbreak{}rows} & rows with at least $\max(50, 0.1n)$ nonzeros & proxy \\
\midrule
\multicolumn{3}{@{}l}{\textbf{blocks} (4): does the instance split into blocks?} \\
\texttt{components}, \texttt{largest\_\allowbreak{}component\_\allowbreak{}cols\_\allowbreak{}frac} & connected components of the row--column graph & exact \\
\texttt{components\_\allowbreak{}without\_\allowbreak{}dense\_\allowbreak{}rows}, \texttt{largest\_\allowbreak{}component\_\allowbreak{}without\_\allowbreak{}dense\_\allowbreak{}rows\_\allowbreak{}cols\_\allowbreak{}frac} & the same after removing dense rows & proxy \\
\midrule
\multicolumn{3}{@{}l}{\textbf{symmetry} (9): are there duplicate or interchangeable columns?} \\
\texttt{duplicate\_\allowbreak{}cols}, \texttt{duplicate\_\allowbreak{}col\_\allowbreak{}groups}, \texttt{duplicate\_\allowbreak{}rows} & identical columns and rows & exact \\
\texttt{refined\_\allowbreak{}col\_\allowbreak{}classes}, \texttt{refined\_\allowbreak{}row\_\allowbreak{}classes}, \texttt{largest\_\allowbreak{}col\_\allowbreak{}class}, \texttt{cols\_\allowbreak{}in\_\allowbreak{}shared\_\allowbreak{}classes\_\allowbreak{}frac} & classes from colour refinement; only columns in a shared class can be interchangeable & sufficient \\
\texttt{refinement\_\allowbreak{}rounds}, \texttt{refinement\_\allowbreak{}stable} & rounds of refinement; whether it converged & exact \\
\bottomrule
\end{tabularx}
\caption{Structure-profile fields. Provenance: exact, sufficient (asserted only when proven), proxy (fixed threshold, or computed after removing the proxy dense rows). The role of each group is given in \cref{tab:structure_roles}.}
\label{tab:structure_fields}
\end{table*}

\clearpage
\section{What the Structure Profile Can Suggest}
\label{app:structure}
We choose each field of the structure profile for what it can tell the agent about the instance: a field earns its place if it can suggest a hypothesis about a bottleneck, or rule out a direction in the search space (\cref{tab:search-space}).
\Cref{tab:structure_roles} gives this role group by group.
Two further conditions apply to every field: it is computed by reading the instance file without solving it, and it holds only counts, fractions, and orders of magnitude, so that it reveals neither names nor values.

\paragraph{Suggest or rule out.}
The two roles do not carry the same weight, and the provenance label says which role a field can play.
A field labelled exact or sufficient can rule a direction out: no indicator constraints means that \texttt{LinearizeIndicators} has nothing to act on, and a change to it is expected to be a no-op change.
A proxy field, or a field that states only a necessary condition, can only suggest a hypothesis, which the next log must then test.
The skill states this asymmetry explicitly: the structure profile qualifies a bottleneck or rules a direction out, but it is not evidence that a bottleneck exists (\cref{sec:method:interp}).

\begin{table*}[t]
\centering
\footnotesize
\setlength{\tabcolsep}{3pt}
\renewcommand{\arraystretch}{1.05}
\begin{tabularx}{\textwidth}{@{}l>{\raggedright\arraybackslash\hsize=1.15\hsize}X>{\raggedright\arraybackslash\hsize=0.95\hsize}X>{\raggedright\arraybackslash\hsize=0.9\hsize}X@{}}
\toprule
\textbf{Group} & \textbf{What it can suggest} & \textbf{Directions it bears on} & \textbf{What it cannot show} \\
\midrule
size & scale of the LP; with the \emph{presolve} group of the log profile, how much presolve removed & \texttt{LpMethod} and barrier controls for large LPs; \texttt{Presolve} & difficulty; size is context, not evidence \\
\midrule
variables & pure binary or mixed instance; general integers and their domain widths; implied integers, which let the objective step be proven; free and unbounded variables & directions that target general integers are ruled out when there are none; \texttt{Presolve}, \texttt{LpMethod}, \texttt{Dualize} & implied integers are found only through equality rows within five rounds; bounds may be tightened by presolve \\
\midrule
objective & a zero objective makes the instance a feasibility problem; a proven objective step converts gaps into steps (\cref{app:fields}) & with a zero objective, only the primal side matters: heuristic levels; a dual segment with a few steps left points to cuts, branching, and conflict analysis rather than heuristics & \texttt{objective\_integral}=false means ``not proven'', not ``not integral'' \\
\midrule
constraints & the dominant row class describes the formulation: set partitioning, packing, and covering rows give clique and conflict structure; knapsack rows admit cover cuts; mixed-binary and variable-bound rows suggest on--off structure; continuous rows put the cost in the LP & \texttt{Presolve}, cut levels, and \texttt{StrongBranching} for set rows; cut levels for knapsack rows; LP controls for continuous rows; \texttt{LinearizeIndicators} and \texttt{LinearizeSos} are ruled out when the counts are zero & covers linear rows only; the classes say how the formulation is written, not how COPT uses it \\
\midrule
numerics & big-M rows suggest a weak LP relaxation and numerical risk; a large row dynamism suggests numerically difficult LPs & cut levels and \texttt{Presolve}, read together with the numerical-issue fields of the log profile; \texttt{LpMethod}, \texttt{DualPerturb}, \texttt{DualPrice}, \texttt{BarHomogeneous} & the big-M threshold is a convention; a root that already closes most of the gap contradicts a weak relaxation; the search space has no tolerance or scaling parameter \\
\midrule
sparsity & dense rows and columns make each LP iteration and barrier factorisation costly; dense rows are the coupling rows removed in \emph{blocks} & \texttt{LpMethod}, \texttt{BarOrder}, \texttt{Dualize}; tree cut levels when cuts on dense rows are costly & the dense-row threshold is a convention \\
\midrule
blocks & several components, or several after removing dense rows, suggest independent or loosely coupled subproblems; one component that holds almost all columns rules separability out & no parameter controls decomposition; \texttt{Presolve} and sub-MIP heuristics only as hypotheses & whether COPT exploits blocks is not visible in the log; coupling columns are not detected \\
\midrule
symmetry & duplicate columns prove that a symmetry exists; columns alone in their colour class cannot be interchanged by any symmetry; shared classes only allow symmetry & \texttt{Folding}$^\dagger$ and \texttt{Presolve}; ruled out when every class is a single column & shared classes do not prove symmetry; symmetry created by presolve is not seen \\
\bottomrule
\end{tabularx}
\caption{Role of each group of the structure profile (\cref{tab:structure_fields}). Directions name parameters of the search space (\cref{tab:search-space}); $\dagger$ marks an internal control whose levels are not publicly documented. A direction is a hypothesis for the next evaluation, never a rule.}
\label{tab:structure_roles}
\end{table*}

\paragraph{Objective step.}
The objective step is the field through which the structure profile enters the log profile.
When every column with a nonzero objective coefficient is integral or implied integral, and every coefficient becomes integral after scaling by $10^k$ with $k\le 6$, the difference of any two feasible objective values is a multiple of the step, the greatest common divisor of the scaled coefficients divided by $10^k$.
The log profile then reports gaps in steps (the \texttt{\_steps} fields of \cref{tab:log_fields_a,tab:log_fields_b,tab:log_fields_c}), so the agent can tell a dual segment with a few steps left from one with thousands; a gap below one step means that no better solution exists.
When the step cannot be proven, the log profile assumes a unit step only if the printed incumbents are integral, and labels the objective step as assumed.
Implied integers follow the presolve reasoning of \citet{achterberg2020presolve}: a continuous column is integral if it is the only continuous column of an equality row whose coefficients and right-hand side, divided by its coefficient, are integral.

\paragraph{Row classes.}
Each linear row falls into the first of 17 classes whose rule it satisfies: empty, free, singleton; set partitioning, packing, covering; cardinality, invariant knapsack; aggregation, precedence, variable bound; equation knapsack, knapsack, integer knapsack; mixed binary, continuous; and general.
The classes are adapted from MIPLIB 2017 \citep{MIPLIB2017}, but the order and details differ, so the fractions cannot be compared directly with MIPLIB statistics.
The agent uses them to describe the formulation in one sentence, for example ``a pure binary instance dominated by set-packing rows, with an integral objective of step~1''.

\paragraph{Why colour refinement can rule symmetry out.}
A symmetry of the instance is a pair of permutations of rows and columns that preserves column types, bounds, objective coefficients, row sides, and every matrix coefficient.
Colour refinement \citep{grohe2014dimension} starts from colours that every symmetry preserves, columns by type, bounds, and objective coefficient and rows by their sides, and in each round recolours every row and column by its colour and the multiset of neighbour colours and coefficients.
Each new colour is a function of data that every symmetry preserves, so by induction every symmetry maps each column to a column of the same colour, after every round.
A column alone in its class is therefore fixed by every symmetry, and if every class is a single column the instance file has no column-permutation symmetry.
This conclusion holds even when refinement stops at its round or time limit, and even under hash collisions, because both only make the classes coarser; in that case \texttt{cols\_in\_shared\_classes\_frac} is an upper bound.
The converse fails: columns that share a class need not be interchangeable, and regular structure produces large classes without any symmetry.
Shared classes therefore only suggest that symmetry handling \citep{pfetsch2019computational} or \texttt{Folding} is worth testing.

\begin{figure}[htbp]
\centering
\includegraphics[width=\textwidth]{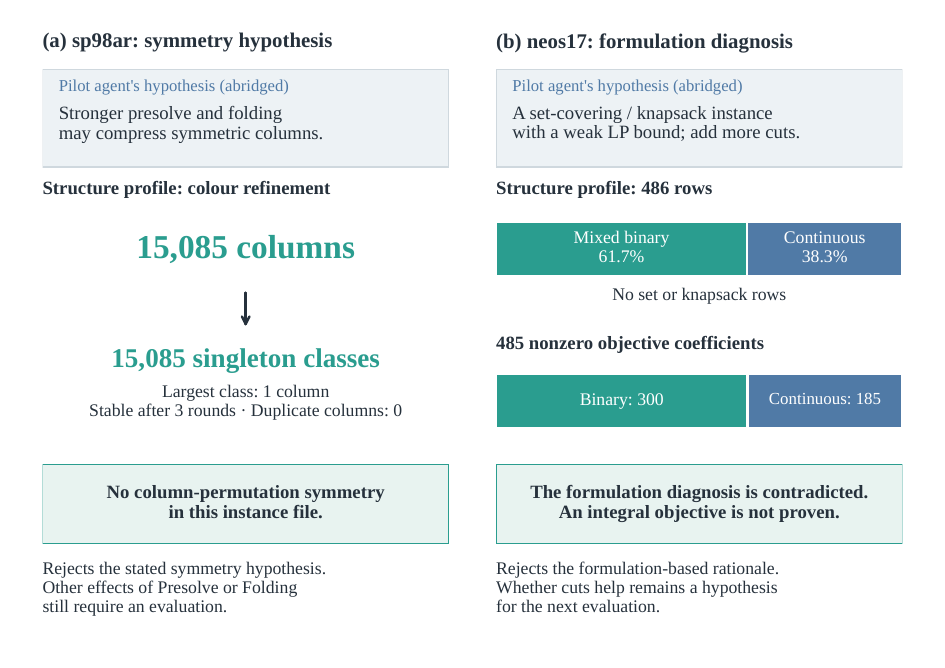}
\caption{Structure evidence can reject a hypothesis or its premise before an evaluation. The pilot agent's hypotheses are abridged from its recorded rationales; the structure profiles were computed afterwards. (a) Every column of \texttt{sp98ar} lies in a singleton colour class, ruling out column-permutation symmetry in the instance file. (b) \texttt{neos17} has mixed-binary and continuous rows, with no set or knapsack rows; its objective is not proven integral. These observations reject the stated premises, without establishing the effects of presolve, folding, or cuts.}
\label{fig:structure-evidence}
\end{figure}

\paragraph{Two cases from the trajectory study.}
The pilot agent of \cref{app:study} had no structure profile; we computed the structure profile afterwards from the same instance files.
On \texttt{sp98ar}, the pilot agent set \texttt{Presolve}=3 and \texttt{Folding}=3 on the hypothesis that stronger presolve and folding would compress a large set of symmetric columns; the change left the node count unchanged on two of the five permuted copies and was not faster than the default configuration.
In the structure profile, refinement is stable after three rounds with all 15{,}085 columns in classes of their own, and no column is duplicated: the instance file has no column-permutation symmetry, and the hypothesis could have been ruled out before the evaluation.
On \texttt{neos17}, the pilot agent described the instance as a set-covering or knapsack problem with a weak LP bound and planned more cuts.
In the structure profile, 61.7\% of the rows are mixed binary and 38.3\% continuous, with no set or knapsack row, and 185 of the 485 objective coefficients lie on continuous columns, so the objective is not provably integral: the description that framed the plan was wrong.
Both cases show what the structure profile can rule out; whether it improves the speedup is not tested in this paper.
\Cref{fig:structure-evidence} pairs the hypotheses with the corresponding evidence, computed from V1 of each instance.
The absence of column-permutation symmetry rules out the stated symmetry hypothesis on \texttt{sp98ar}, while the row classes contradict the formulation diagnosis on \texttt{neos17}; neither conclusion establishes which configuration will be faster.

\paragraph{Limits.}
The structure profile describes the instance file before presolve, and COPT may change any of its fields.
The log does not show whether COPT exploits a block structure or a symmetry, so a direction that the structure profile suggests is confirmed or refuted only by the next log.

\clearpage
\section{What the Log Cannot Explain}
\label{app:unexplained}
\method aims at tuning whose changes can be explained by the log, but not every change in solving time can be.
This appendix lists six sources of changes that the log cannot trace to a mechanism, and how \method treats them.
They are not failures of reading: they remain when every field is read correctly.

\paragraph{Noise.}
Two runs with an identical printed search path can still differ in solving time.
On \texttt{anonymous\_33} (\cref{app:case:anonymous}), \grimip evaluated its selected configuration ten times, and the solving times range from 61.7 to 67.5\,s.
The base comparison identifies this case: an identical printed search path whose time difference is spread over the whole run counts as noise, and the promotion margin of 2\% keeps small differences of this kind from replacing the incumbent configuration.

\paragraph{Performance variability.}
A change can alter the printed search path for reasons unrelated to its mechanism: a different order of operations or a different tie-break sends the tree search elsewhere, and the solving time changes with it.
MIP solvers are known to vary in this way under changes that leave the instance mathematically unchanged, such as a permutation of rows and columns or a different random seed \citep{lodi2013performance}.
The log shows that the printed search path diverged and where, but not whether the divergence came from the mechanism that the hypothesis names or from performance variability; a single run cannot separate the two.

\paragraph{Mechanisms the log does not print.}
The default COPT log omits cut, heuristic, and restart statistics (\cref{tab:log_availability}).
A change to these mechanisms can change the solving time while the fields that describe them stay unobserved; the base comparison then finds a time difference concentrated in one interval, which points to a change that the log does not print.
On \texttt{anonymous\_33}, the agent read a node cost of 12.0\,ms after aggressive heuristics as heuristic overhead; the log shows the node cost, but the heuristic time behind it is an inference that only the next change can test.

\paragraph{Effects that cross phases.}
A bottleneck describes the observed cost in one phase; a change in an earlier phase can alter that cost.
An upstream hypothesis therefore compares both the earlier phase and the subsequent search, even when the earlier phase accounts for little solving time.
In the trajectory study (\cref{app:study}), a change reduced the node count from 286 to 162 and increased the solving time from 13.03 to 16.14\,s: the prediction on the node count held, and the solving time moved the other way for reasons the log does not attribute.

\paragraph{Joint changes.}
When a change sets several parameters, or one parameter that controls several mechanisms, the log shows the joint effect but not its attribution.
An interaction hypothesis tests such a combination relative to a stated base configuration; comparisons with individual changes are needed to separate their contributions.
\texttt{CutLevel} sets root and tree cuts together; on \texttt{anonymous\_33} only two further controlled parameter changes showed that the gain came from the tree cuts and that the root cuts had hidden part of it.
The configuration selected by SMAC on the same instance switches off cuts, presolve, and strong branching together, and its log cannot say which of the three produced the speedup.

\paragraph{Truncated runs and proxies.}
A truncated run gives only one-sided values, and phase times are proxies.
The diagnosis therefore stops at the deepest level that the provenance and comparability labels support (\cref{sec:method:interp}); below that level, a difference between two runs has no reading.

\paragraph{The unexplained verdict.}
\method does not force an explanation onto these changes.
Besides confirmed and refuted, a verdict can be \emph{unexplained}: the solving time changed beyond what the base comparison counts as noise, but the new log profile cannot trace the change to the fields named in the prediction.
An unexplained verdict does not block the candidate: promotion and final selection depend on solving time and final gap (\cref{sec:method:system}), so the selected configuration may carry a gain that the log does not explain.
The verdict is intended to prevent attribution of that gain to the hypothesis, and reuse of the hypothesis as if it had been confirmed; the audit in \cref{app:verdict-audit} checks how well the recorded reasoning follows this distinction.
Because each candidate is evaluated once, \method cannot tell an unexplained gain caused by performance variability from one caused by a mechanism the log does not print; repeated runs on permuted copies of the instance would separate them, at a cost in budget.
\paragraph{Observed frequency and final selection.}
In the verdict audit (\cref{app:verdict-audit}), unexplained accounts for 173/2,505 linked primary labels (6.9\%) and occurs in 126/339 trajectories with evaluations (37.2\%).
Of the 33 competing labels, 31 also include unexplained; counting all competing labels gives 204/2,538 labels (8.0\%).
These are frequencies of the agent's recorded wording: some uses explicitly describe noise or final-gap differences, so they do not all satisfy the narrower definition above.

We assess final selection by following the \emph{base chain}, from the selected configuration through its recorded base configurations to Default.
An observed gain means a speedup of at least 1.02 relative to a solved base, solving an unsolved base, or a strictly smaller final gap when both runs reach the full time limit.
Among the 274 non-default selections in the sessions with the recording protocol, 27 (9.9\%) directly select such a gain labeled unexplained, and 56 (20.4\%) have one anywhere in their base chain.
The latter corresponds to 56/358 (15.6\%) when default selections are included.
These counts measure recorded dependence through the base chain, not whether an inherited parameter is causally necessary for the final gain.
A further 52/274 base chains have missing, competing, or unlinked verdicts without an identified unexplained gain; the last evaluation has no following rationale.
The reported counts therefore establish the presence of such selections within this recording channel, without classifying all remaining gains as explained.

\section{Search Space}
\label{app:space}
\Cref{tab:search-space} lists the 42 COPT parameters available to \method, organized by the solver operations they control.
A configuration specifies non-default values; omitted parameters retain their default values.
Integer domains include every integer between the listed endpoints, and real domains are continuous.
The value $-1$ denotes automatic control where specified by COPT; its meaning is parameter-dependent.
Some parameters are internal controls accepted by COPT~8.1 but lack public documentation of their individual levels.
We mark these parameters in the table and do not assign undocumented levels additional mechanistic meanings.

The search space is narrower than the solver's full set of legal values: \texttt{Presolve}=4 is excluded, and \texttt{LpMethod} is restricted to $\{-1,0,1,2\}$.
Run controls, including thread count, time limit, and logging, are fixed by \cref{sec:exp:setup} and are not tunable; the relative-gap tolerance retains its solver default.
Parameters outside the table also retain their default values.

Before an evaluation, the system layer checks parameter types, domains, and coherence.
For example, it rejects non-default barrier controls with dual simplex (\texttt{LpMethod}=1), restart controls when restarts are disabled, and presolve controls when presolve is disabled.
The frozen implementation used for the main runs also rejects simultaneous non-default assignments to \texttt{CutLevel} and its root or tree counterparts, and individual heuristic levels when \texttt{HeurLevel}$\in\{0,1,2\}$.
These additional rules enforce a single parameter representation in the agent interface.
The baseline search spaces are specified in \cref{app:baselines}.

\begin{table*}[t]
\centering
\footnotesize
\setlength{\tabcolsep}{2pt}
\begin{minipage}[t]{0.49\textwidth}
\begin{tabular*}{\linewidth}{@{}l@{\hspace{3pt}\extracolsep{\fill}}ccc@{}}
\toprule
\textbf{Parameter} & \textbf{Type} & \textbf{Def.} & \textbf{Domain} \\
\midrule
\multicolumn{4}{@{}l}{\textit{Presolve}} \\
\texttt{Presolve} & int & $-1$ & $[-1,3]$ \\
\texttt{PresolveRestart}$^\dagger$ & int & $-1$ & $[-1,3]$ \\
\texttt{Folding}$^\dagger$ & int & $-1$ & $[-1,3]$ \\
\midrule
\multicolumn{4}{@{}l}{\textit{LP}} \\
\texttt{LpMethod} & int & $-1$ & $\{-1,0,1,2\}$ \\
\texttt{Dualize} & int & $-1$ & $[-1,1]$ \\
\texttt{DualPrice} & int & $-1$ & $[-1,1]$ \\
\texttt{DualPerturb} & int & $-1$ & $[-1,1]$ \\
\midrule
\multicolumn{4}{@{}l}{\textit{Cuts}} \\
\texttt{CutLevel} & int & $-1$ & $[-1,3]$ \\
\texttt{RootCutLevel} & int & $-1$ & $[-1,3]$ \\
\texttt{TreeCutLevel} & int & $-1$ & $[-1,3]$ \\
\texttt{RootCutRounds} & int & $-1$ & $[-1,M_{\mathrm I}]$ \\
\midrule
\multicolumn{4}{@{}l}{\textit{Heuristics}} \\
\texttt{HeurLevel} & int & $-1$ & $[-1,3]$ \\
\texttt{RoundingHeurLevel} & int & $-1$ & $[-1,3]$ \\
\texttt{DivingHeurLevel} & int & $-1$ & $[-1,3]$ \\
\texttt{FapHeurLevel} & int & $-1$ & $[-1,4]$ \\
\texttt{SubMipHeurLevel} & int & $-1$ & $[-1,3]$ \\
\texttt{Heur57}$^\dagger$ & int & $-1$ & $[-1,3]$ \\
\texttt{Heur71}$^\dagger$ & int & $-1$ & $[-1,3]$ \\
\midrule
\multicolumn{4}{@{}l}{\textit{Sub-MIP/LNS}} \\
\texttt{SubMipParallel}$^\dagger$ & int & $1$ & $[0,2]$ \\
\texttt{SubMipSetting}$^\dagger$ & int & $0$ & $[0,1]$ \\
\texttt{HeurLnsDisableFAPFac}$^\dagger$ & real & $0$ & $[0,1]$ \\
\texttt{HeurLnsDisableLNSFac}$^\dagger$ & real & $0$ & $[0,1]$ \\
\bottomrule
\end{tabular*}
\end{minipage}
\hfill
\begin{minipage}[t]{0.49\textwidth}
\begin{tabular*}{\linewidth}{@{}l@{\hspace{3pt}\extracolsep{\fill}}ccc@{}}
\toprule
\textbf{Parameter} & \textbf{Type} & \textbf{Def.} & \textbf{Domain} \\
\midrule
\multicolumn{4}{@{}l}{\textit{Branching/conflict}} \\
\texttt{ConflictAnalysis} & int & $-1$ & $[-1,1]$ \\
\texttt{StrongBranching} & int & $-1$ & $[-1,3]$ \\
\texttt{SubStrongBranching}$^\dagger$ & int & $-1$ & $[-1,3]$ \\
\midrule
\multicolumn{4}{@{}l}{\textit{Restarts}} \\
\texttt{RestartFreq}$^\dagger$ & real & $1$ & $[10^{-4},10^3]$ \\
\texttt{RestartKeepRootCuts}$^\dagger$ & int & $0$ & $[0,1]$ \\
\texttt{MaxRestarts}$^\dagger$ & int & $M_{\mathrm I}$ & $[0,M_{\mathrm I}]$ \\
\midrule
\multicolumn{4}{@{}l}{\textit{Reformulation}} \\
\texttt{LinearizeIndicators} & int & $0$ & $[0,1]$ \\
\texttt{LinearizeSos} & int & $0$ & $[0,1]$ \\
\texttt{ReformBinProds}$^\dagger$ & int & $1$ & $[0,1]$ \\
\texttt{ReformBinProdsWithBin}$^\dagger$ & int & $0$ & $[0,1]$ \\
\midrule
\multicolumn{4}{@{}l}{\textit{Analytic center}} \\
\texttt{AnalyticCenterEffort}$^\dagger$ & int & $1$ & $[0,3]$ \\
\texttt{AnalyticCenterWorkLimit}$^\dagger$ & real & $-1$ & $[-1,M_{\mathrm R}]$ \\
\midrule
\multicolumn{4}{@{}l}{\textit{Barrier}} \\
\texttt{BarHomogeneous} & int & $-1$ & $[-1,1]$ \\
\texttt{BarOrder} & int & $-1$ & $[-1,2]$ \\
\texttt{BarStart} & int & $-1$ & $[-1,2]$ \\
\texttt{BarReduceSupport}$^\dagger$ & int & $-1$ & $[-1,1]$ \\
\texttt{BarRegMethod}$^\dagger$ & int & $-1$ & $[-1,1]$ \\
\texttt{BarLinSystemMethod}$^\dagger$ & int & $-1$ & $[-1,1]$ \\
\midrule
\multicolumn{4}{@{}l}{\textit{Other strategies}} \\
\texttt{RestoreLpDiving}$^\dagger$ & int & $1$ & $[0,1]$ \\
\texttt{SolveRootSocRelax}$^\dagger$ & int & $1$ & $[0,1]$ \\
\bottomrule
\end{tabular*}
\end{minipage}
\caption{Search space of \method, transcribed from the experiment configuration.
Here $M_{\mathrm I}=2{,}147{,}483{,}647$ and $M_{\mathrm R}=1.7976931348623\times10^{308}$.
Def.: default value. Intervals are inclusive and interpreted according to the type column.
$\dagger$ marks internal controls whose level semantics are not fully documented publicly.
\grimip uses the same parameter names with three upper-bound changes described in \cref{app:baselines}.}
\label{tab:search-space}
\end{table*}

\section{Baselines}
\label{app:baselines}
The baselines address instance-specific tuning: each instance is tuned independently, without sharing evaluation histories across instances.
They use the run controls in \cref{sec:exp:setup} and the same cached default run.
All tuners receive budget $B=10\,t(c_0)$; SMAC and \grimip allow at most 50 candidate evaluations, excluding the cached default run.
An evaluation starts only when at least $0.5\,t(c_0)$ remains.
The search procedures and their treatment of truncated runs are detailed below.

\subsection{Default}
Default selects $c_0$ without tuning.
Its solving time and final gap come from the cached default run, which also supplies $t(c_0)$ to all tuners.

\subsection{SMAC}
We use the algorithm configuration facade of SMAC3~2.3.1~\citep{lindauer2022smac3}.
The fixed search space follows the six Gurobi parameters of SMAC in \citet{luo2026grimip}: \texttt{MIPFocus}, \texttt{Heuristics}, \texttt{Cuts}, \texttt{Presolve}, \texttt{Method}, and \texttt{VarBranch}.
We map them to their COPT counterparts \texttt{HeurLevel}, \texttt{CutLevel}, \texttt{Presolve}, \texttt{LpMethod}, and \texttt{StrongBranching}.
COPT has no counterpart of \texttt{MIPFocus}, so \texttt{RestartFreq} takes the sixth position.
The domains are those in \cref{tab:search-space}; all other parameters retain their default values.
The integer parameters are represented as categorical levels, and \texttt{RestartFreq} is continuous on a logarithmic scale.

The initial design contains only $c_0$, whose cost is supplied from the cache without consuming budget.
SMAC uses a random forest with 10 trees, expected improvement, and local and sorted random search to select candidates.
Random candidates are interleaved with probability $0.5$, and the forest is refitted after each new evaluation.
The scenario is deterministic, with search seed 0 and at most one run per configuration on the instance; previously evaluated configurations reuse their recorded results.
For non-failed runs, the cost supplied to SMAC is $\log_2(t(c)/t(c_0))$, which has the same ranking as solving time on the fixed instance.

Each run is subject to $\min(2t(c_0),1800\,\mathrm{s})$ and the remaining budget.
Unlike \method, SMAC does not tighten this cap using the incumbent configuration.
For a truncated run, it receives the log-ratio cost evaluated at the cap, a lower bound on the completed-run cost.
SMAC's own intensifier determines its incumbent configuration during search, without the promotion margin of \method.
Final selection follows the common rule in \cref{sec:method:system}: when Default solves the instance, an eligible candidate must achieve speedup at least $1.02$; when Default reaches the time limit, a candidate must solve the instance or reach the full time limit with a strictly smaller final gap to replace $c_0$.
Truncated and failed runs are excluded, and ties retain $c_0$.

\subsection{Random}
Random samples candidate configurations at random for each instance from the same six-dimensional search space as SMAC: \texttt{HeurLevel}, \texttt{CutLevel}, \texttt{Presolve}, \texttt{RestartFreq}, \texttt{LpMethod}, and \texttt{StrongBranching}.
The parameter domains are those in \cref{tab:search-space}, and all other parameters retain their default values.
Candidate generation uses neither a surrogate model nor an LLM.
Each candidate is evaluated on the current instance under the common run settings and budget $B=10\,t(c_0)$ (\cref{sec:exp:setup}).

\subsection{\grimip}
We port the LLM-driven BO procedure of \citet{luo2026grimip} to COPT, retaining automated search-space selection, warm-starting, and expected improvement.
Before warm-starting, the LLM selects at most six parameters from the pool in \cref{tab:search-space}; this subset is fixed throughout tuning.
For this baseline, the upper bounds of \texttt{RootCutRounds}, \texttt{MaxRestarts}, and \texttt{AnalyticCenterWorkLimit} are reduced to 100; the remaining domains are unchanged.
Unspecified parameters retain their solver defaults, including defaults outside these restricted search intervals.

The initial information comprises static instance features and dynamic features extracted from progress rows within the first 30\,s of the cached default run; no separate preliminary run is performed.
Static features describe variable and constraint counts, variable types, nonzeros, and matrix density.
Dynamic features describe the available root, incumbent, bound, and node progress.
Quantities absent from COPT logs, such as cut counts and maximum tree depth, are omitted.
Following the original procedure, \grimip also receives the instance name and file name.
It receives no log or profile from subsequent evaluations: the BO history contains only configurations and scalar solving times.

Warm-starting generates five initial candidates.
At each subsequent round, the LLM generates three candidates, predicts a mean and standard deviation for each, and selects the candidate with the largest expected improvement for evaluation.
The LLM is DeepSeek-V4.1-Flash with thinking disabled, temperature $0.1$, top-$p$ $0.8$, and a maximum of 8,000 output tokens per call.
If direct uncertainty prediction fails, the implementation falls back to five sampled performance predictions.
The 50-evaluation limit includes the initial candidates.
Patience-based early stopping is disabled; tuning ends when the budget no longer permits an evaluation, the evaluation limit is reached, or candidate generation fails.

The budget charges solver wall time and the wall time of each LLM phase; parallel LLM calls within a phase are charged by their elapsed wall time.
The default run is not charged.
The evaluation cap is
\[
  \min\!\left(2t(c_0),\,2t_{\mathrm{inc}},\,1800\,\mathrm{s},\,B_{\mathrm{rem}}\right),
\]
where $B_{\mathrm{rem}}$ is the remaining budget and $t_{\mathrm{inc}}$ is the smaller of the default time and the fastest completed \grimip run.
This cap uses every strict improvement, without the promotion margin of \method.
Following the original BO feedback convention, a truncated run enters the BO history with its observed solving time, without a penalty or a truncation indicator; the cached default result is not inserted into that history.
Final selection follows the common rule in \cref{sec:method:system}, as described for SMAC above.
As for the other tuners, the selected configuration is not run again.

\section{Prompts}
\label{app:prompts}
This section lists the prompt texts used in the recorded tuning sessions of \method, in the order in which they appear.
The raw-log variant omits the profile summaries and skill catalog and loads no skill; the system rules and search space remain available.
The system prompt (\cref{app:prompts:system}) states the rules of the system layer and the search space; the kickoff message (\cref{app:prompts:kickoff}) is appended to it and describes the instance, the default run, and the budget.
The tools (\cref{app:prompts:tools}) are available for the whole session.
The first user message starts the session and lists the skill, which the agent loads on demand (\cref{app:prompts:control,app:prompts:skill}).
All texts except the skill are quoted verbatim from the code; placeholders in braces are filled per instance.
The instruction below for final selection when the default run reaches the time limit is historical; reported results apply the default-preserving final selection in \cref{sec:method:system} to the recorded evaluations.
The texts use the terms of this paper; identifiers keep their code names: \texttt{log\_features} and \texttt{model\_profile} are the log and structure profiles, and the default configuration also answers to \texttt{"baseline"}.
Source: \texttt{src/prompts.py} and \texttt{src/single.py} in the tuning-agent repository, exported by \texttt{docs/export\_paper\_prompts.py}.

\subsection{System Prompt}
\label{app:prompts:system}
\texttt{\{space\}} is the description of the 42 parameters of \cref{tab:search-space}, one line per parameter with its type, range, and default value.
The budget, cap, and promotion values are those of \cref{sec:method:system}.
\begin{lstlisting}[style=prompt]
You are a COPT MIP parameter-tuning agent. Tune one instance in one session. Use evaluate for solver runs, read_log for de-identified logs, list_files/read_file/grep_files for the read-only workspace. The system controls session termination and final selection.

## Objective and selection
speedup = t_default / t (higher is better; times below 0.01 s count as 0.01 s). A time-limit run counts at the limit.
When the default run is solved, a complete candidate qualifies if its speedup is at least 1.02; the system strictly selects the highest speedup, otherwise the default configuration.
When the default run times out, select the fastest solved candidate; if none solves, select the smallest final gap among candidates reaching the full time limit. Missing gap counts as infinity. There is no default fallback when an eligible candidate exists, even if its gap is worse than default. Runs stopped at a shorter cap, failed runs and budget-interrupted runs cannot be selected. Exact ties between improving candidates use evaluation order.

## Evaluations and budget
One run per configuration. Build a candidate as base + changes; bases are "default", "incumbent", or an evaluation id. Reset a parameter with "default". Duplicate configurations are rejected.
The cap is min(2 x default time, 2 x the incumbent configuration's time) and at most the time limit 1800 s. A run stopped at a shorter cap is truncated, cannot win and is not run again.
The incumbent configuration starts at default. With a solved default, promotion requires speedup >= 1.02 x the incumbent configuration's speedup. With a timed-out default, promotion compares solved time first, then final gap. An incumbent alone means a solver solution, not a configuration.
Solver, LLM and shell wall time consume the budget. Keep testing while the budget permits. Only the system ends the session when the remaining budget cannot fund another evaluation or model step. You cannot end tuning early; if a direction is exhausted, formulate a new hypothesis and continue.

## Experiments and explanations
Read the default log before evaluating. Compare each candidate with its base: state the observed bottleneck, a mechanism hypothesis and predicted changes in named log quantities. After observing the result, assign the hypothesis a verdict: confirmed if those changes appear, refuted if contradicted, unexplained if the evidence is insufficient to attribute the outcome. Include the previous verdict in the next rationale.
A single run does not establish causality. Search-path changes can alter time substantially; a time gain alone does not confirm a mechanism. Test a close variation before building on a surprising gain. Compare truncated runs at matched progress rather than treating their terminal nodes or gap as completed results. A combination of parameter changes does not isolate each parameter's contribution.

## Parameters
The tool enforces consistency (e.g. barrier controls with LpMethod=1, restart controls with MaxRestarts=0, a group switched off globally while one of its members is raised). One evaluation may change multiple parameters or functional groups (presolve, LP method, cuts, heuristics, branching/conflict, restarts, ...).
Know how settings combine:
   - Mixed CutLevel and RootCutLevel/TreeCutLevel settings are allowed. The coherence policy rejects CutLevel=0 with a positive RootCutLevel or TreeCutLevel.
   - Individual heuristic levels are allowed with HeurLevel=1 or 2. The coherence policy rejects HeurLevel=0 with a positive individual heuristic level.
   - Do not infer which setting takes precedence or whether a change has no effect from the parameter names alone; compare the observed runs.
   - An explicit level can behave differently from automatic (-1). To remove a parameter, reset it with "default".
Use round values for continuous parameters (e.g. RestartFreq 0.1, 0.3, 0.5, 2, 5, 10, 20, 50 — not 0.33 or 45).

## Workspace
/session.json contains the default run and selection rules; /evaluations.json indexes all runs. Each /configs/<config>/ contains params.json, evaluation.json, V1/result.json and V1/log.txt (V1 = the instance's model file). Use /README.md for file details.

## Tunable parameters (all other COPT parameters are fixed)
{space}

Call evaluate at most once per turn, using the previous result. Before tool calls, briefly state the evidence and next action.
\end{lstlisting}

\subsection{Kickoff Message}
\label{app:prompts:kickoff}
\texttt{\{features\}} gives the solver version, the thread count, and the size of the instance; \texttt{\{summary\}} is a summary of the default run's log of at most 4000 characters, made of the header lines, a sample of progress rows, and the footer.
\texttt{\{log\_profile\_short\_view\}} and \texttt{\{structure\_profile\_short\_view\}} are the short views of the log profile of the default run and of the structure profile, with one line per group.
\begin{lstlisting}[style=prompt]
## Instance (identity withheld)
{features}

## Default run
Default run: {default_time} s {default_state}{default_nodes}
Full log: read_log(run="default").

## Budget
{budget} s (= 10 x the default time {default_time} s).
Read the default log, then evaluate from base="default". Evaluation ids start at E1.

Load skill(name="copt-log-profile") before interpreting profiles. Full log_features and base_comparison are in each configuration's V1/result.json; model_profile is in /session.json. Short views follow.
{log_profile_short_view}
{structure_profile_short_view}
Full structure profile (model_profile): /session.json.

Shell: the bash tool runs each command in a fresh sandbox. The workspace is mounted read-only at /workspace (its files also at /, as read_file names them, e.g. /configs/default/V1/result.json); /scratch is writable and kept between commands. python3 has the standard library only; there is no network and no solver: evaluate remains the only way to run the solver. A command stops after 60 s unless it asks for up to 120 s (timeoutMs). Its wall time is charged to the budget, and a command whose timeout exceeds the budget left is rejected. The skills' own files (references, scripts) are readable in the sandbox under the directory the skill tool names.
\end{lstlisting}

\subsection{Tools}
\label{app:prompts:tools}
The tool name, its arguments, the description, and the descriptions of the arguments.
\texttt{skill} and \texttt{bash} come from the agent framework; we abbreviate the generic parts of the \texttt{bash} description with ``[\dots]''.
Each \texttt{evaluate} reply states the result, the cap, the budget left, and the short view of the new log profile.
\begin{lstlisting}[style=prompt]
evaluate(base, changes, rationale)
  Run one candidate (`base` + `changes`); return time, gap, speedup and promotion outcome. Consumes budget. Invalid or duplicate configurations are rejected without a solver run.
    base: Base configuration: "default" (the default configuration; "baseline" is accepted too), "incumbent" (the incumbent configuration) or an evaluation id such as "E3".
    changes: Parameter -> new value. Use "default" to reset a parameter to its COPT default. Parameters not listed keep their value from `base`.
    rationale: Previous hypothesis verdict (confirmed / refuted / unexplained) with evidence, then the current mechanism hypothesis and predicted log changes.

read_log(run, start, max_lines, grep)
  Read the de-identified COPT log of one run: "default" (the default run) or an evaluation id. Returns numbered lines; page with `start`/`max_lines` or filter with `grep` (case-insensitive regular expression). File names and paths are removed; all solver values are shown as printed. Costs only your own time.
    start: first line (1-based); negative = count from the end

list_files(path, depth)
  List your workspace (read-only): the directory of every evaluated configuration with its parameters, evaluation, run result and log. Paths are relative to the workspace root "/". Costs only your own time.
    path: directory or file, e.g. "/" or "/configs" (default "/")
    depth: levels to expand, 1-4 (default 1)

read_file(path, start, max_lines, grep)
  Read one workspace file with numbered lines, e.g. /evaluations.json, /configs/<config>/evaluation.json, /configs/<config>/V1/result.json or /configs/<config>/V1/log.txt. Page with `start`/`max_lines` or filter with `grep` (case-insensitive regular expression). Costs only your own time.
    start: first line (1-based); negative = count from the end

grep_files(pattern, path, name)
  Search the workspace files for a case-insensitive regular expression and return matching lines as path:line: text — e.g. compare one line of the log across all configurations with pattern="Solve node", name="log.txt". Costs only your own time.
    path: directory or file to search (default "/")
    name: file-name glob, e.g. "log.txt" or "*.json"

skill(name)
  Load the full instructions for a skill. Call it before acting on a task that names or clearly matches a skill in the session skill catalog.

bash(command, description, timeoutMs, workdir)
  Execute a bash command (`bash -c`) and return its stdout/stderr. Each call runs in a fresh shell; pass `workdir` instead of using `cd`. [...] Long output is truncated to its tail; the full output is saved to a file whose path is reported when available. [...]
\end{lstlisting}

\subsection{Session Messages}
\label{app:prompts:control}
The first user message carries the skill catalog, where \texttt{\{skill\_description\}} is the description line of the skill (\cref{app:prompts:skill}).
A nudge follows a reply without a tool call.
When the context exceeds 800k tokens, the agent writes notes under the compaction instruction, and the system appends its exact record of all evaluations as \texttt{\{state\}}.
\begin{lstlisting}[style=prompt]
[First user message]
Begin the tuning session: the instructions, the instance and the default run are in the system prompt.

<system-reminder>
A skill is a reusable set of task-specific instructions. The following skills are available in this session:

<available_skills>
- `copt-log-profile`: {skill_description}
</available_skills>

If the user names a skill, or the task clearly matches a skill's description, call the `skill` tool with the exact skill name before taking task actions. Load all applicable skills, then follow their full instructions. This catalog contains summaries only; do not infer or follow a skill's instructions until it has been loaded.
[...]
</system-reminder>

[Nudge, after a reply without a tool call]
Continue by calling one of the tools (evaluate, read_log, list_files, read_file, grep_files). Write your short analysis in the message, then make the call.

[Compaction instruction]
Summarize working notes in at most 200 words: the observed bottleneck; tested hypotheses with verdicts (confirmed / refuted / unexplained) and evidence; remaining uncertainty; next experiment. The system appends exact evaluation records, so do not duplicate them. Reply with notes only, without tools.

[Appended to the compaction summary]
## Exact record kept by the tuning system (at compaction)
{state}
\end{lstlisting}

\subsection{Skill}
\label{app:prompts:skill}
The skill \texttt{copt-log-profile} consists of \texttt{SKILL.md}, two references that document the fields of the log and structure profiles with their labels (\cref{app:fields}), the two profiling scripts, and a log comparison script.
\texttt{SKILL.md} is written in Chinese; below is our English translation in the terms of this paper, with field names and the diagnostic steps preserved.
Step~3 describes the observed cost without restricting the parameter search to its phase; Step~4 asks the agent to choose one hypothesis of type direct, upstream, or interaction for the next evaluation.
These are short names for Direct intervention, Upstream intervention, and Interaction intervention in \texttt{SKILL.md}.
\begin{lstlisting}[style=prompt]
---
name: copt-log-profile
description: Interpret COPT MIP structure profiles, log profiles and run comparisons to locate bottlenecks and design testable parameter experiments. For COPT instance-specific tuning: use before reading any run result, writing a rationale or proposing a new configuration. Not for Gurobi logs.
---

# COPT log diagnosis

Before each new configuration, complete the 4 steps below in order and check the checklist at the end before submitting. When evidence is insufficient, describe only the supported observations, state the unknown causes, and then generate a hypothesis to test.

## Step 1: Read the profiles (do not recompute)
The host has computed the profiles; read them directly. Do not recompute them from masked logs.
- Structure: model_profile in /session.json. Fields: references/model-profile.md.
- One run: log_features in /configs/<config>/V1/result.json. The default run is /configs/default/.
- Comparison with the base configuration's raw log: base_comparison in the same result.json.
- Short views: the kickoff and evaluate returns.
Field provenance and comparability under truncation: references/features.md; read the relevant group when needed. Resolve ambiguity in the raw log, checking the header and phase first. Profiles omit absolute objective values. Visible logs may mask objectives and bounds as obj; unprinted value changes and instance identity cannot be recovered, do not try.

Standalone (no host profiles): python3 scripts/copt_model_profile.py <model> only reads the model (requires coptpy, numpy, scipy); python3 scripts/copt_log_features.py <log> --objective-step <step> parses the log; python3 scripts/copt_log_compare.py --help gives comparison usage. Log scripts use the standard library only; --summary gives a short view.

## Step 2: Check data availability
Before descending, you must check:
1. Host state / lower_bound / failed.
2. The completeness, timing quality and value availability of the profile's run group.
Structure describes the formulation and limits hypotheses; it does not replace run evidence.

## Step 3: Diagnose the observed cost
Describe what happens in the dominant time segment, level by level. Where time is spent does not restrict the parameter search: most time spent in the tree neither locates the cause in the tree nor limits candidates to tree parameters. Earlier presolve and root work, and interactions between parameters, can change the subsequent search.
This step asks where the time goes and what happens during it; keep observations distinct from causes. When evidence ends, stop at the current level and name the missing fields. Leave causes for Step 4 as hypotheses; do not infer a parameter group directly from a phase label.
1. L1: presolve / root / tree. Use phases timing proxies to find the dominant segment. Boundaries come from printed rows; missing boundaries prevent a split. Truncated durations are lower bounds.
2. L2: describe progress in that segment.
   - Presolve: examine reductions and time together.
   - Root: distinguish the root LP, other rounds and silent time, and inspect the root bound.
   - Tree: split at the last incumbent update into primal / dual segment. Primal segment: distinguish no incumbent from slow improvement; it also advances the bound, so its duration is not all solution-search time. Dual segment: distinguish slow bound movement from exhausting open nodes.
3. L3 (tree only): describe node volume and node cost using nodes, tree_ms_per_node, lpit_per_node, windows and bound progress. Only a completed run reveals the total node volume required for that search. Time per node includes mixed overhead; LPit/n is a printed proxy. Neither is pure node-LP CPU time.

## Step 4: Generate cross-phase hypotheses
For the observations in Step 3, consider the three types below and choose one hypothesis for this evaluation. Choose using existing comparisons, possible causes not yet covered, expected cost, and predictions that can distinguish explanations; do not choose by whether a parameter belongs to the dominant phase.
- direct (Direct intervention): change work within the observed phase. For high tree node cost, compare tree-cut or strong-branching overhead with pruning gains.
- upstream (Upstream intervention): change an earlier phase to test its effect on later cost. For a slow tree, compare stronger, weaker or disabled presolve and the resulting formulation and search; or change root cuts and compare subsequent tree size and bound progress. A short earlier phase can strongly affect a later one.
- interaction (Interaction intervention): jointly change interacting parameters to test whether the combination changes the current trade-off. A cut configuration may help only with particular presolve or heuristic configurations. State the base configuration, evaluate the joint change, and compare with existing individual changes when needed. A combination's gain cannot be assigned to one parameter.

The rationale names the chosen type, base configuration, changes, expected link between the intervention and the observations, and log changes that would support or refute it. For a cross-phase hypothesis, inspect both the earlier phase and later time, nodes, or primal/bound progress.
An unsuccessful strengthening of a parameter constrains only the tested configuration and context; it does not rule out weakening, disabling, or combining it with other changes. More presolve reduction does not automatically imply less total solving time. One effective local combination does not rule out gains elsewhere in the search space.
Choose one type per evaluation; the three types help generate candidates and are not a fixed sequence of required evaluations. Parameter directions remain hypotheses to test, not diagnostic conclusions.
When comparing LP options, distinguish the initial LP from subsequent reoptimization; lower time per node alone does not prove faster node LPs. Structure may support more specific directions, such as Folding for symmetry evidence; absent indicator/SOS constraints leave their linearization controls without a target, so do not tune them. Current system rules define allowed parameters and consistency.

## Interpretation limits (check each before stating any conclusion)
- Incumbent and gap: the first finite incumbent may appear in an ordinary row before the first H; H/* may repeat the same solution. A smaller gap alone does not show bound movement. obj, -- and infinity are not numerical evidence. Printed gap=0 does not establish completion of the proof.
- Root rounds: windows ending at H/non-H rows mix work; they are not precise cut/heuristic times. Missing cut counts, restart counts and similar quantities remain unknown; do not fill them in.
- Comparison with the base: equal signature means equal recorded columns and alone does not show an ineffective parameter. Check base_comparison's first difference and timing distribution: an equal printed path with diffuse timing changes supports noise; changes concentrated in one phase may still reflect unprinted work. A different path does not by itself establish the proposed mechanism.
- Truncated runs: align at equal time, nodes or gap; base_comparison.matched_progress gives shared progress. Node totals and average node cost at different endpoints are not directly comparable. Inspect primal/bound progress to determine which side changes the gap.

## Output: observations and hypothesis in the rationale
Report the observation path, key evidence and stopping level in one sentence, then state the chosen hypothesis:
Observation: <L1> -> <L2> -> <L3 or reason for stopping>; evidence: <key evidence>
Hypothesis: <direct / upstream / interaction>; base=<configuration>; <intervention and expected link>; prediction: <observations that support or refute it>
Example: Observation: tree -> dual segment -> high node volume; evidence: the final incumbent appears early, with many nodes processed afterwards. Hypothesis: upstream; base=current configuration; weaker presolve may change the subsequent search; prediction: compare the presolved formulation, tree node volume and total solving time to test whether extra tree work offsets changes in earlier work or node cost.
Follow the system prompt for hypotheses, predictions, the three verdicts and final selection. Do not assign a combination's gain to one component or build on unexplained acceleration as an established mechanism.

## Pre-submit checklist
- [ ] Used the host profiles; did not recompute from masked logs.
- [ ] Checked state/lower_bound/failed and the run group's completeness, timing quality and value availability.
- [ ] Descended level by level; stopped where evidence ends and named the missing fields.
- [ ] Step 3 describes observations without restricting the parameter search to the phase where time is spent.
- [ ] Step 4 selects one type (direct / upstream / interaction) and states the base configuration, intervention link and testable prediction.
- [ ] Wrote parameter directions as hypotheses to test; did not rule out weakening, disabling or combinations after one unsuccessful strengthening.
- [ ] Did not take obj, --, infinity, gap=0 or a smaller gap alone as evidence.
- [ ] Aligned truncated runs at shared progress; did not compare node totals or average node cost at different endpoints.
- [ ] Did not judge a parameter ineffective from equal signature alone, nor a mechanism established from a different path alone.
- [ ] The rationale contains an observation sentence (path + evidence + stopping level) and the chosen hypothesis.
- [ ] Did not assign a combination's gain to one component or build on unexplained acceleration as an established mechanism.
\end{lstlisting}

\section{More Related Work}
\label{app:related}

\paragraph{Profiling-guided LLM agents.}
LLM agents that write GPU kernels use execution and profiling feedback~\citep{ouyang2025kernelbench}; recent systems place a profiler in the loop to guide diagnosis and subsequent changes~\citep{zhang2025cudaforge,lei2025pragma}.
This practice motivates the separation between deterministic log profiling and the agent's interpretation in \method.

\paragraph{Explaining configurations.}
Functional ANOVA quantifies the contribution of individual parameters to performance variation~\citep{hutter2014fanova}, while ablation analysis attributes the performance difference between two configurations to parameter changes~\citep{fawcett2016ablation}.
These methods explain performance across configurations using additional analysis.
\method records log evidence and a hypothesis during tuning, then checks the prediction against a subsequent evaluation.
The two forms of explanation are complementary: parameter attribution can help assess the agent's account of a change.

\paragraph{Learning inside the solver.}
Learned branching~\citep{gasse2019gcnn}, node selection~\citep{he2014learning}, and primal heuristics~\citep{nair2020solving} guide internal solver decisions.
The ML4CO competition~\citep{gasse2022ml4co} studies such policies in a common evaluation framework, alongside broader work on learning for combinatorial optimization~\citep{bengio2021ml4co}.
\method operates through exposed solver parameters and logs; learning internal decisions addresses a different level of solver control.

%% file: sections/miplib_trajectories.tex
\subsection{Additional MIPLIB Trajectories}
\label{app:miplib-trajectories}
We extend the evaluation view in \cref{fig:case-study}(a) with 20 trajectories from the frozen MIPLIB results underlying \cref{tab:main}.
For display, we rank instances with completed default and selected runs, a default time of at least 10\,s, and at least five evaluations by speedup, and show the first 20.
These examples are selected for large observed improvements; they do not estimate typical performance or explanation correctness.
Every panel retains the complete evaluation sequence, including slower candidates and truncated runs.
Annotations identify actual changes relative to the stated base configuration; they do not establish a causal mechanism.
In grouped labels, root/tree denotes \texttt{RootCutLevel}/\texttt{TreeCutLevel}; all listed assignments are applied together.

\begin{figure}[!htbp]
\centering
\includegraphics[width=0.49\textwidth]{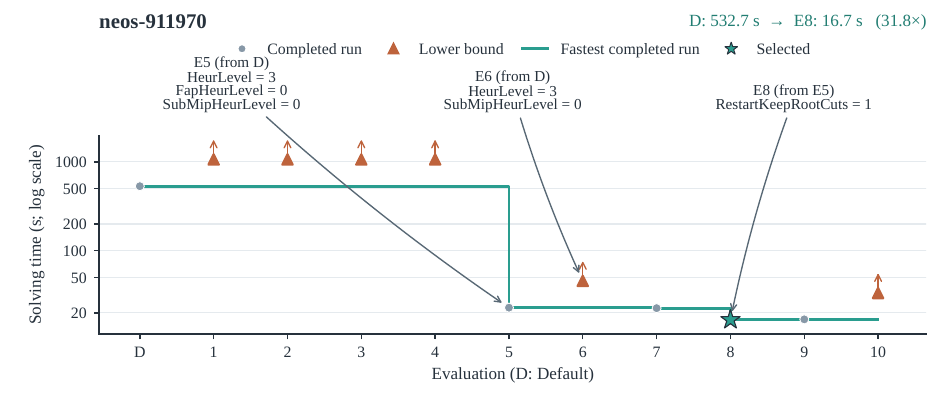}\hfill
\includegraphics[width=0.49\textwidth]{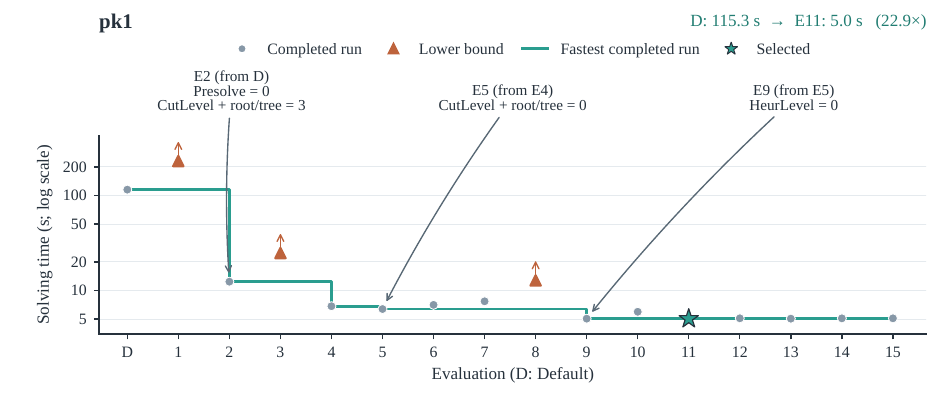}\par\smallskip
\includegraphics[width=0.49\textwidth]{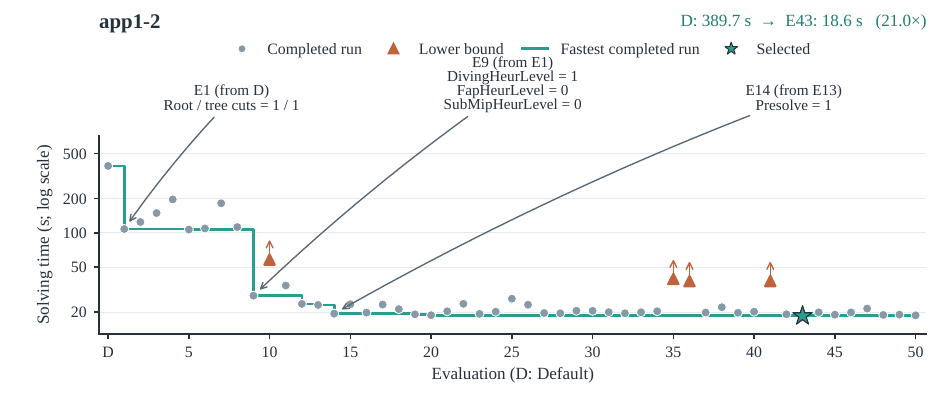}\hfill
\includegraphics[width=0.49\textwidth]{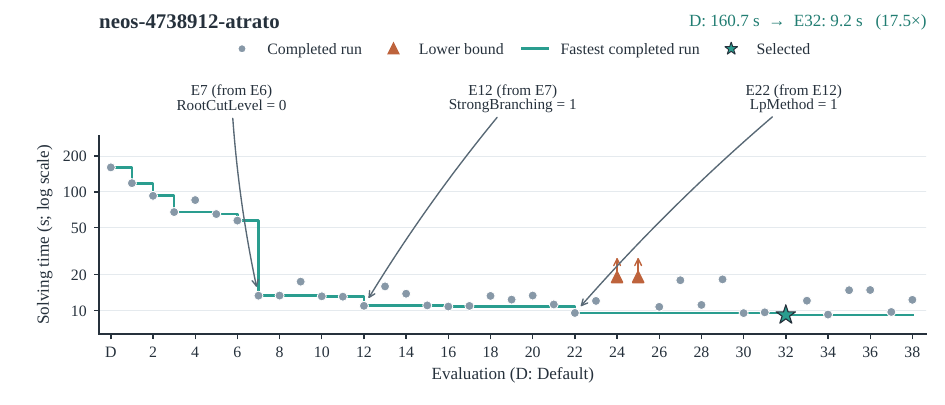}\par\smallskip
\includegraphics[width=0.49\textwidth]{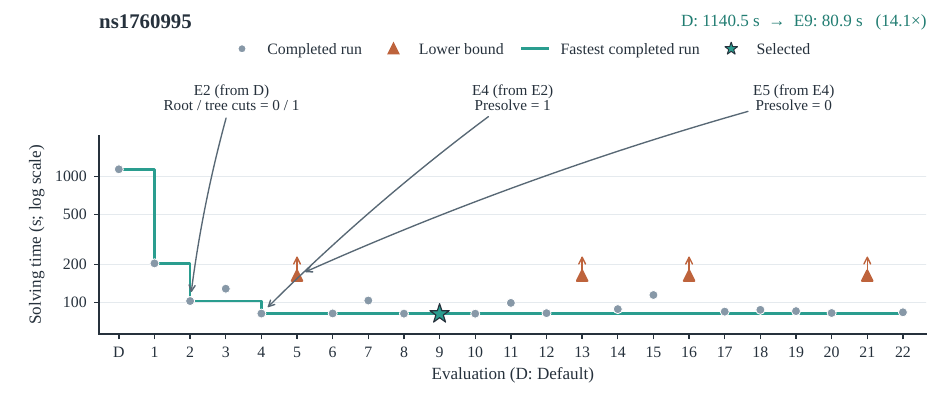}\hfill
\includegraphics[width=0.49\textwidth]{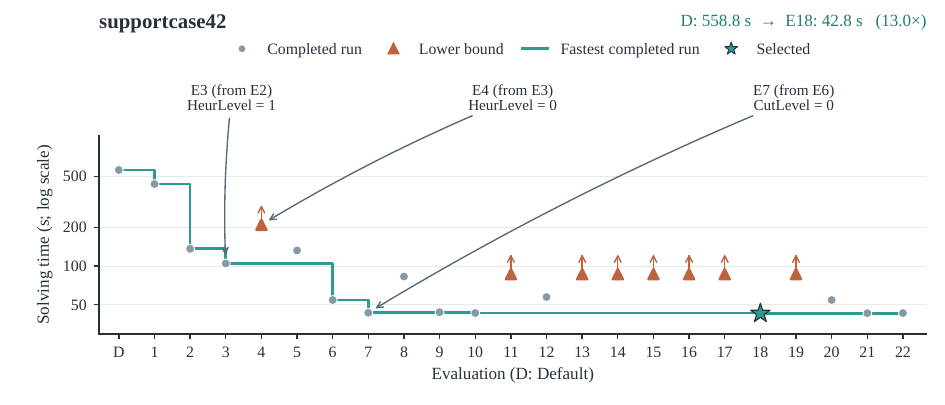}\par\smallskip
\includegraphics[width=0.49\textwidth]{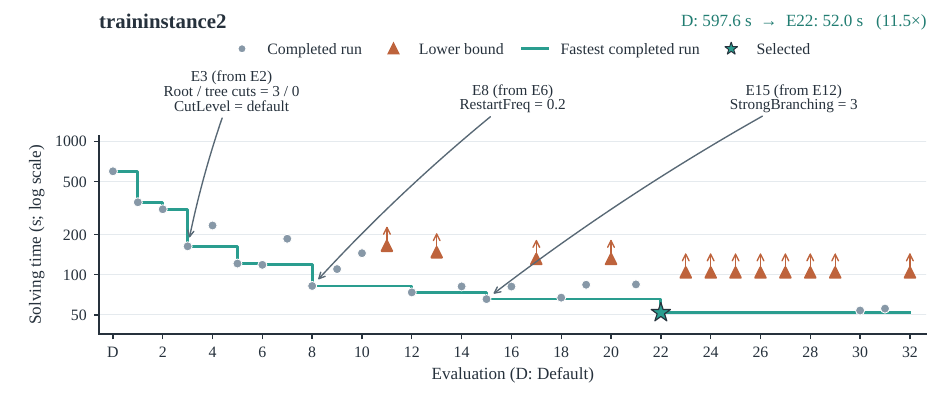}\hfill
\includegraphics[width=0.49\textwidth]{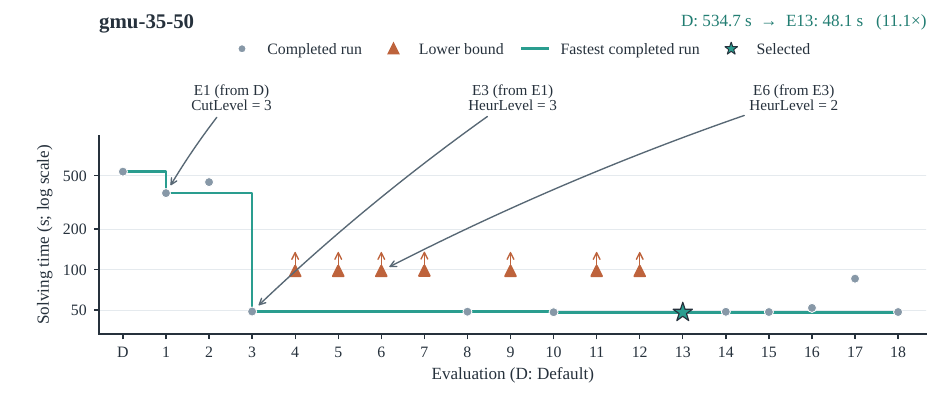}\par\smallskip
\includegraphics[width=0.49\textwidth]{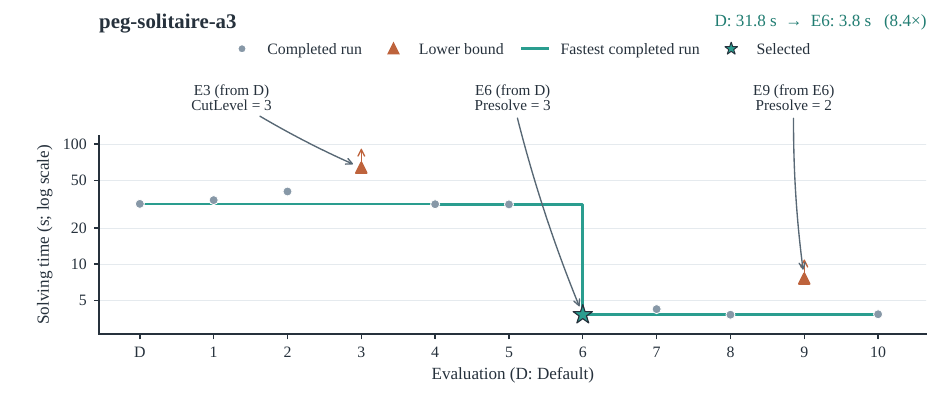}\hfill
\includegraphics[width=0.49\textwidth]{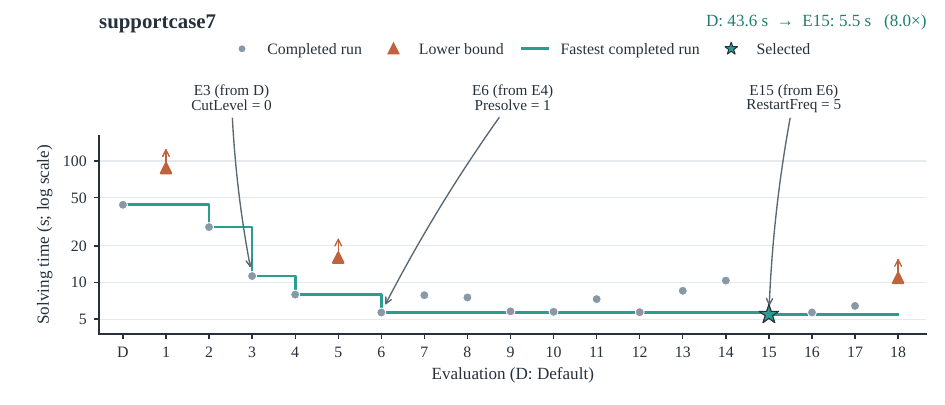}
\caption{Additional MIPLIB trajectories (1/2), read left to right. D denotes Default. Circles mark completed runs; upward triangles and arrows mark lower bounds on completed solving times for truncated runs. The solid line follows the fastest completed run, including Default; the star marks the selected configuration. Each annotation names its evaluation and base configuration. Solving time uses a logarithmic axis, with a separate range for each instance.}
\label{fig:miplib-trajectories-1}
\end{figure}
\clearpage

\begin{figure}[p]
\centering
\includegraphics[width=0.49\textwidth]{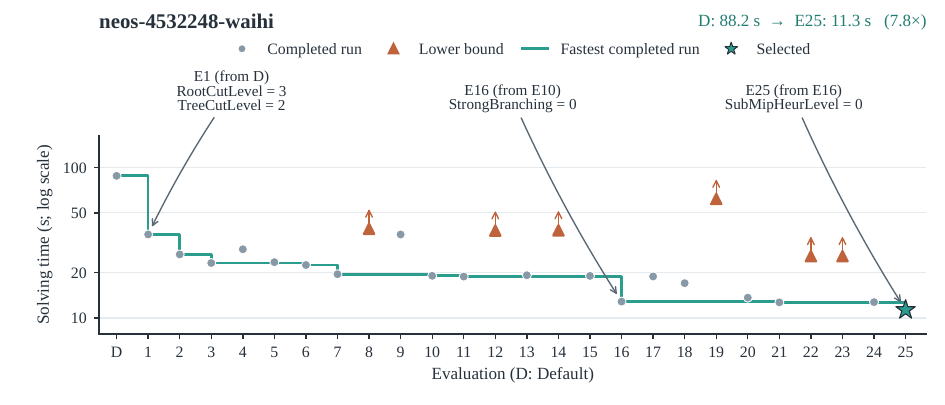}\hfill
\includegraphics[width=0.49\textwidth]{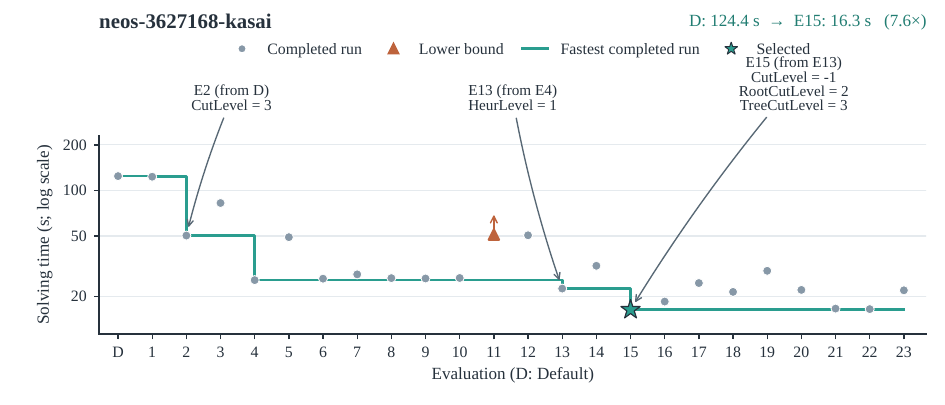}\par\smallskip
\includegraphics[width=0.49\textwidth]{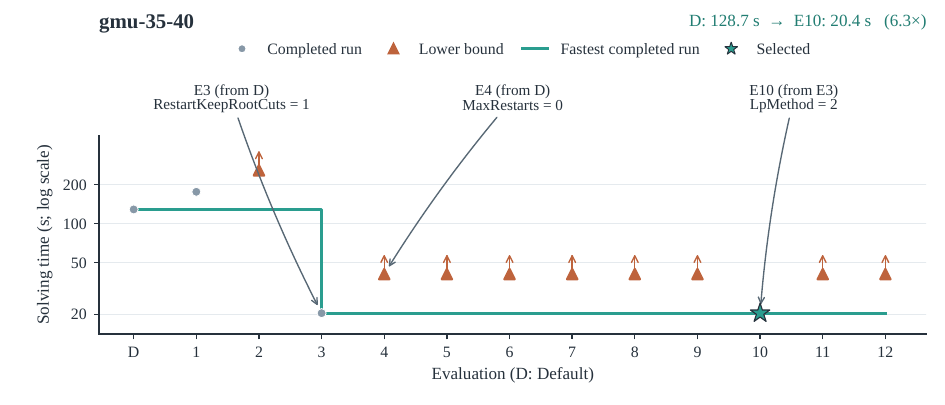}\hfill
\includegraphics[width=0.49\textwidth]{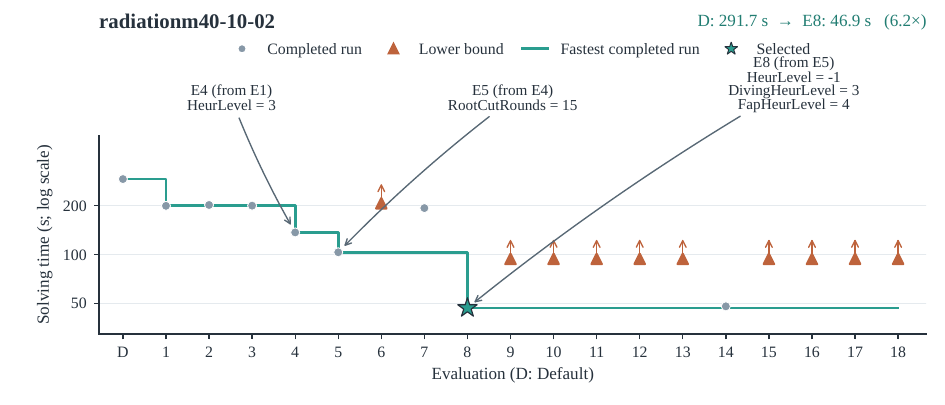}\par\smallskip
\includegraphics[width=0.49\textwidth]{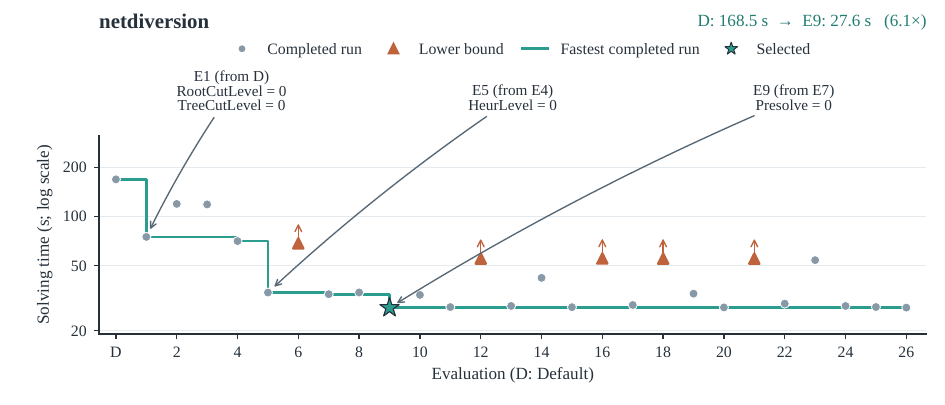}\hfill
\includegraphics[width=0.49\textwidth]{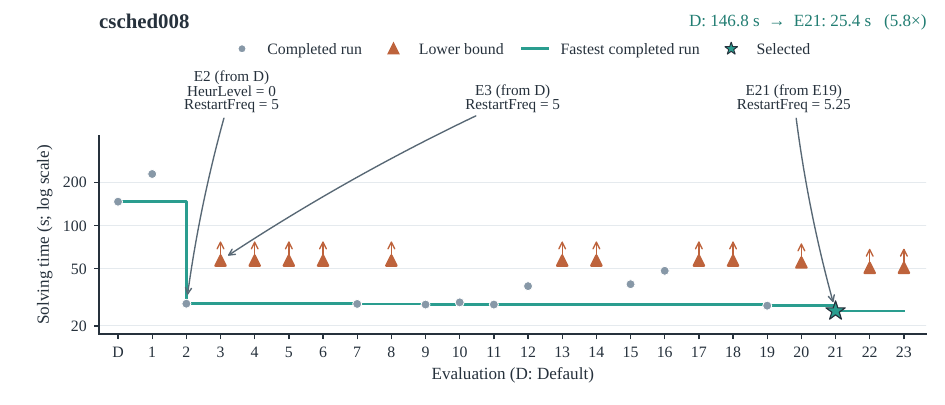}\par\smallskip
\includegraphics[width=0.49\textwidth]{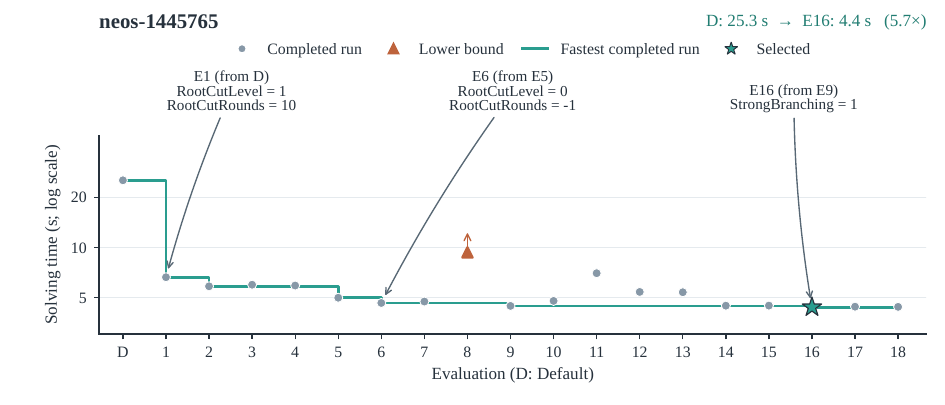}\hfill
\includegraphics[width=0.49\textwidth]{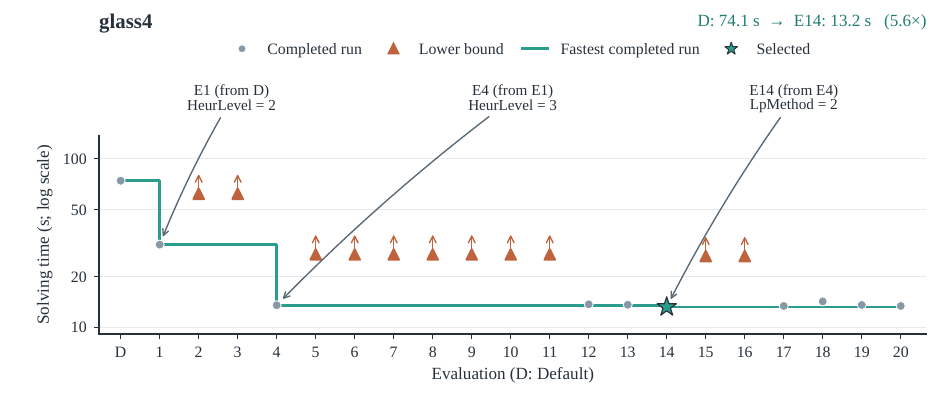}\par\smallskip
\includegraphics[width=0.49\textwidth]{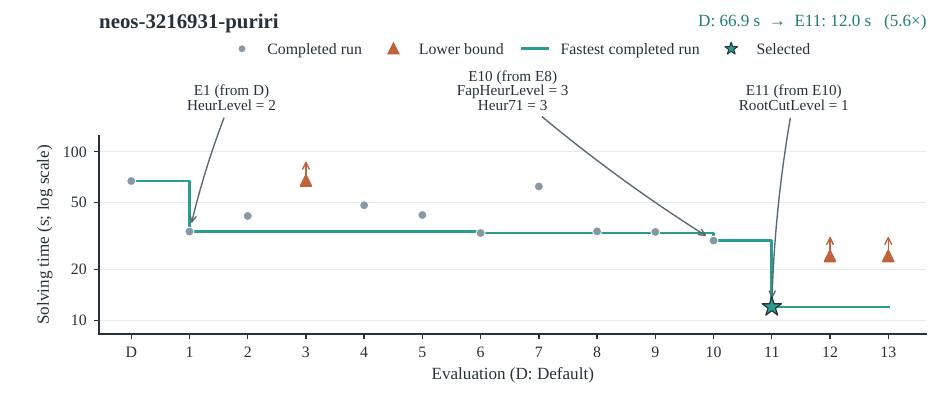}\hfill
\includegraphics[width=0.49\textwidth]{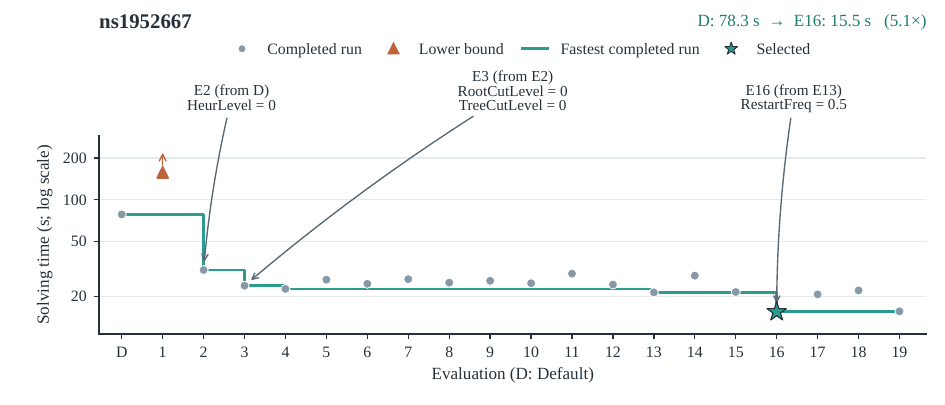}
\caption{Additional MIPLIB trajectories (2/2), read left to right. All evaluations are shown; notation follows \cref{fig:miplib-trajectories-1}.}
\label{fig:miplib-trajectories-2}
\end{figure}